\documentclass[default,iicol,sn-aps]{sn-jnl}

\usepackage{graphicx}%
\usepackage{multirow}%
\usepackage{amsmath,amssymb,amsfonts}%
\usepackage{amsthm}%
\usepackage{mathrsfs}%
\usepackage[title]{appendix}%
\usepackage{xcolor}%
\usepackage{textcomp}%
\usepackage{manyfoot}%
\usepackage{booktabs}%
\usepackage{algorithm}%
\usepackage{algorithmicx}%
\usepackage{algpseudocode}%
\usepackage{listings}%

\usepackage{mathtools}%
\usepackage{comment}%
\usepackage[caption=false, font=footnotesize]{subfig}%
\usepackage{tikz}%
\usepackage{pgfplots}%
\pgfplotsset{compat=1.18}%
\begin{document}

\title[Article Title]{\centering Centralized vs. Decoupled Dual-Arm Planning\\Taking into Account Path Quality}

\author*[1]{\fnm{Jonas} \sur{Wittmann}}\email{jonas.wittmann@tum.de}

\author[1]{\fnm{Franziska} \sur{Ochsenfarth}}\email{franziska.ochsenfarth@tum.de}

\author[1]{\fnm{Valentin} \sur{Sonneville}}\email{valentin.sonneville@tum.de}

\author[1]{\fnm{Daniel} \sur{Rixen}}\email{rixen@tum.de}

\affil[1]{Technical University of Munich, TUM School of Engineering and Design, Department of Mechanical Engineering, Chair of Applied Mechanics; Munich Institute of Robotics and Machine Intelligence (MIRMI); \orgaddress{\street{Boltzmannstr. 15}, \city{Garching}, \postcode{85748}, \country{Germany}}}

%
\abstract{The aim of coordinated planning is to avoid robot-to-robot collisions in a multi-robot system, and there are two standard solution approaches: centralized planning and decoupled planning. Our first contribution is a decoupled planning approach that ensures $\mathcal{C}^2$-continuous control commands with zero velocities at the start and goal. We benchmark our decoupled approach with a centralized approach. Contrary to literature, we show that for a standard motion planning pipeline, such as the one used by \textit{MoveIt!}, centralized planning is superior to decoupled planning in dual-arm manipulation: It has a lower computation time and a higher robustness. Our second contribution is an optimization that minimizes the rotational motion of an end-effector while considering obstacle avoidance. We derive the analytic gradients of this optimization problem, making the algorithm suitable for online motion planning. Our optimization extends an existing path quality improvement method. Integrating it into our decoupled approach overcomes its shortcomings and provides a motion planning pipeline that is robust at up to 99.9\,\% with a planning time of less than 1\,s and that computes high-quality paths.}%
%

\maketitle

%
\section{Introduction}\label{sec:Introduction}%
Single-arm manipulation is a state-of-the-art automation solution for many industrial processes, such as welding, and solutions have been proposed for various applications like human-robot collaboration or time-optimal planning. There are ready-to-use frameworks such as \textit{MoveIt!}~\cite{Coleman2014} that implement a complete manipulation process with a simple user interface: The user or vision system defines a target object, and the framework computes collision-free motions, grasping poses, etc. to pick up the object and place it at a defined target pose.\newline%
When it comes to dual-arm manipulation, planning algorithms become more computationally expensive and more complex. Complete framework solutions do not exist. But there is ongoing research~\cite{Nour2023} and current effort concentrates on the implementation of  multi-arm automation solutions in industrial applications: \cite{DualArmMoveit}~aims to implement dual-arm planning in \textit{MoveIt!} and \textit{HYDRABYTE}~\cite{Hydrabyte} is a start-up company that provides software for multi-arm path planning.\newline%
\begin{figure}%
    \centering%
    \includegraphics[width=\columnwidth]{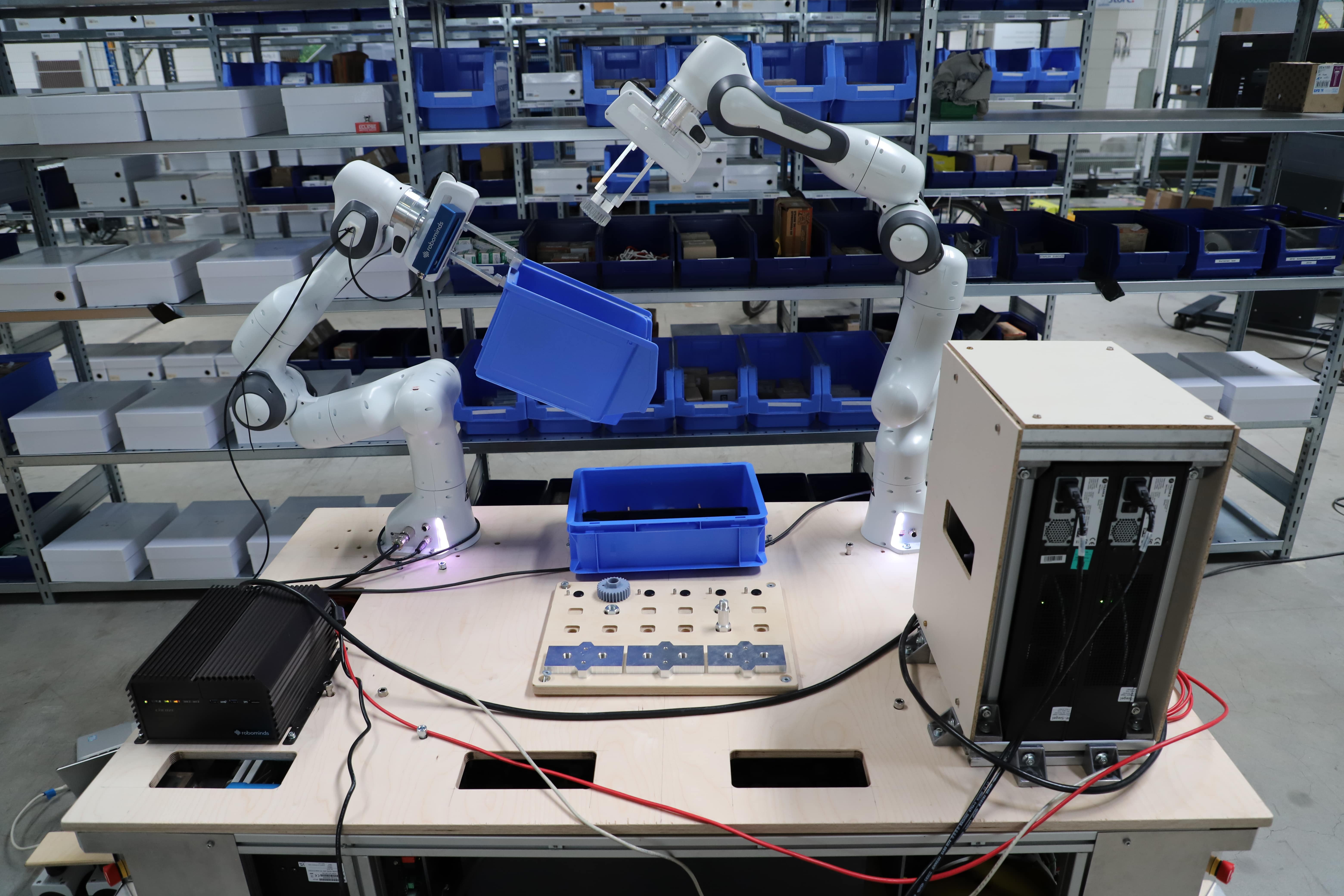}%
    \caption{The hardware of our mobile dual-arm system integrates three components: a custom-built mobile base, two \textit{Franka Emika} robots, and the \textit{robominds} vision system. The camera is mounted on the left end-effector.}\label{fig:DualArmSystem}%
\end{figure}%
There are special algorithms for dual/multi-arm planning, and the reason for this is \textit{scaling}: The planning time of complete single-arm motion planning algorithms, i.e. algorithms that find a collision-free path if one exists, scales exponentially with the number of degrees of freedom (DoF) of the robot~\cite{LaValle2006}. And in addition to the planning algorithm, the computation time of the required collision checking module also scales quadratically with the number of DoFs. Therefore, standard motion planning algorithms may not be suitable for multi-arm systems.\newline%
This work investigates motion planning for dual-arm manipulation. It focuses on \textit{online} planning, i.e. our dual-arm robot computes each incoming motion request before executing it, so the motions are not known/computed before the program launch but during its execution. Figure~\ref{fig:DualArmSystem} shows our mobile dual-arm robot that performs material handling and pre-assembly tasks in a logistics facility.\newline%
Dual-arm planning is a special case of multi-robot planning: It focuses on two robots instead of any number of robots and on robotic manipulators instead of mobile robots. Multi-robot planning is well studied for mobile robots: \cite{Nour2023,Zhi2013}~review path planning for multi-robot formation systems; \cite{Shiwei2022}~presents approaches for aerial, ground, and underwater robots; \cite{Verma2021} studies multi-robot coordination focusing on the navigation problem of mobile robots, but does not consider manipulators. Although there are similarities between motion planning for mobile robots and manipulators, there are also differences that limit the applicability of approaches designed for mobile robots to manipulators. For example, the arrangement of stationary manipulators is predefined, and solution strategies such as swapping two robots as in~\cite{Luna2011} cannot be applied. Therefore, we develop a custom solution.\newline%
The two basic approaches to multi-arm path planning are~\cite{Latombe1991,LaValle2006,Kavraki2008}:%
\begin{enumerate}%
    \item \textit{Centralized} planning: Consider the $n$ robots as a composite robot with $n$ times the DoFs of a single robot, and apply standard motion planning algorithms from single-arm planning to the composite robot.%
    \item \textit{Decoupled} planning: First, plan the motion for each robot separately, considering only collisions with the environment but ignoring other robots. Second, in a post-processing step, coordinate the $n$ computed trajectories so they do not collide.%
\end{enumerate}%
These two approaches differ in terms of \textit{completeness} and \textit{computation time}. Centralized planning is considered computationally inefficient due to the higher number of DoFs and is therefore not applicable to multi-arm systems~\cite{Bouvier2022}. However, it provides \textit{completeness}, i.e. finds a solution if one exists. Decoupled planning is computationally cheaper and therefore preferred, but is not complete~\cite{LaValle2006,Beuke2018a}. For decoupled planning, the two main post-processing strategies to coordinate the computed paths are \textit{prioritized planning} and \textit{fixed-path coordination}~\cite{Latombe1991,LaValle2006,Kavraki2008}. The former first plans paths for higher-priority robots and treats them as obstacles in the subsequent planning phase for lower-priority robots. That is, lower-priority robots must make detours to bypass the path of higher-priority robots~\cite{Bouvier2022}. Further, prioritized planning requires a subsequent check for the particular error case where a lower-priority robot has reached its destination and is blocking the path of a higher-priority robot. So, prioritized planning quickly loses completeness~\cite{LaValle2006}. Fixed-path coordination, which we describe in Section~\ref{sec:Decoupled}, treats path coordination more carefully~\cite{LaValle2006}, and variants of it have been successfully applied in industry~\cite{Beuke2018a}.\newline%
Our work compares a centralized and a decoupled approach based on fixed-path coordination. We evaluate the approaches on the dual-arm robot in Figure~\ref{fig:DualArmSystem}, taking into account the following characteristics:%
\begin{itemize}%
    \item \textit{Robustness}, i.e. what is the success rate of the planner for incoming motion queries?%
    \item \textit{Computation time} for collision-free trajectories.
    \item \textit{Path quality}, which we consider to be the distance traveled by the end-effectors. There are other performance metrics used in multi-robot planning, such as \textit{flowtime}, which is the sum of the motion time of all the robots. However, our robots work alongside humans, so we aim for intuitive motions. Therefore, we want to
 avoid long motion paths resulting from jerky path planning results.%
\end{itemize}%
Our contributions are:%
\begin{itemize}%
    \item In fixed-path coordination, the interpolation scheme is important for applications on real robots because it defines the order of continuity of the final control commands. \cite{Beuke2019} discusses this topic, but still~\cite{Beuke2019,Beuke2018a} use linear interpolation schemes for the fixed-path coordination. Furthermore, they rely on the trajectory generation approach in~\cite{Kunz2012}, which is only \mbox{$\mathcal{C}^1$-continuous}. We show that this leads to discontinuous control commands. Our contribution consists in proposing a decoupled approach based on cubic spline interpolation that, when combined with our trajectory generation approach in~\cite{Wittmann2022}, ensures \mbox{$\mathcal{C}^2$-continuous} control commands with zero velocity at the start and goal in contrast to linear interpolation.%
    \item We propose a path post-processing approach that minimizes the path length in Cartesian space, i.e. the end-effector's combined translational and rotational motion. We build on our approach proposed in~\cite{Wittmann2022a} and add the theory of how to shorten motions in the rotation group~$SO(3)$.%
    \item We benchmark a centralized and a decoupled approach in a real-world dual-arm manipulation scenario. For a standard motion planning pipeline, i.e. when not using our path post-processing approach, we show that, contrary to literature, the centralized approach is superior to the decoupled approach.%
\end{itemize}%
The rest of this paper is organized as follows. In Section~\ref{sec:StateArt}, we present related work. Section~\ref{sec:Benchmark} describes the application scenario we use to benchmark centralized and decoupled planning. The proposed methods are then presented in Section~\ref{sec:PlanningMethod}. Section~\ref{sec:PLPP} outlines our method for improving path quality. The comparison of the planning methods and the benefit of our method for improving path quality are presented in Section~\ref{sec:Results}. Conclusions follow in Section~\ref{sec:Conclusions}.%
%
%
%
\section{Related Work}\label{sec:StateArt}%
Planning for multiple robots has received no~\cite{Siciliano2010} or only little~\cite{Latombe1991,LaValle1998,Choset2005,Kavraki2008} attention in standard motion planning literature, but is well discussed in research. \cite{Smith2012}~reviews coordinated dual-arm planning approaches and highlights \textit{fixed-path coordination}, a decoupled approach originally published in~\cite{Donnell1989}. First, a feasible trajectory is planned separately for each robot. Each computed trajectory ensures that there are no self-collisions for a robot and no collisions with the environment. However, collisions between the two robots can still occur. Therefore, in a subsequent step, the dynamics of both trajectories are modified so that there are no intersections, i.e. collisions, of the trajectories. That is, the paths that the trajectories follow are not modified, but the velocities and accelerations with which the path is traversed are modified. \cite{Beuke2018a}~applies the fixed-path coordination to an \textit{ABB YuMi} robot in a pick-and-place scenario. Compared to hard-coded coordination, where the robots are not allowed to be in the shared workspace simultaneously, fixed-path coordination reduces the execution time by about 16\,\%. And compared to a single-arm solution, the dual-arm solution requires only about 75\,\% of the execution time.\newline%
The fixed-path coordination modifies the trajectories, i.e. the time information of the path. In contrast, some approaches directly coordinate paths without considering time information. \cite{Solovey2016}~proposes \textit{discrete-RRT (dRRT)}. First, it constructs a roadmap for each robot separately, taking into account collisions of that robot with the environment and with itself, but not collisions with other robots. Second, it combines the roadmaps of all robots with the so-called \textit{tensor graph product}~\cite{AndreasVoelz2017}. For a dual-arm setup, this means that a \textit{coordination graph} combines the two single-arm roadmaps: Its nodes are all possible pairs of node indices from the single-arm roadmaps, and edges between these nodes exist if the robots can move simultaneously between the corresponding waypoints without colliding. The size of this \textit{coordination graph} grows exponentially with the number of single-arm roadmaps, so the authors propose an implicitly represented coordination graph. Third, \cite{Solovey2016}~proposes a discrete version of the \textit{RRT} algorithm, originally proposed for continuous space~\cite{LaValle1998a}, to find a feasible path on the coordination graph of a 60 DoF system. In this third step, only collisions between the robots need to be considered. \cite{Shome2019}~builds on~\cite{Solovey2016} and presents \textit{dRRT*}, an asymptotically optimal extension of \textit{dRRT} that considers different formulations of path length as a path quality metric. The algorithm is applied to a \textit{Motoman SDA10F} to reach in a shelf, and the initial solution time, so the time to span a dRRT, is about 500\,ms. Another approach that considers path quality is presented in~\cite{Cohen2014}. The authors use a heuristic-based planning method for manipulation tasks targeting high-DoF systems, which they evaluate on a dual-arm system. First, for a reduced problem, a manipulation lattice graph is created whose nodes are robot configurations and whose edges are two types of collision-free motion primitives between these configurations: static primitives computed offline and adaptive primitives computed online. Each of those are motions with constant joint velocities and 100\,ms duration. The latter are used to reach any goal state independent of the discretization. Instead of a time-critical optimal solution, time-efficient suboptimal solutions with defined cost bounds are computed using \textit{ARA*}. The solutions to the reduced problem are finally used as a heuristic to solve the actual problem. For the heuristic of \textit{ARA*}, \textit{breadth first search} is applied on position-level, but the orientations are neglected. \cite{AndreasVoelz2017}~follows a similar approach as~\cite{Solovey2016} and combines two dynamic roadmaps (DRMs) in a dual-arm application. A DRM is a pre-computed graph in configuration space that stores, in addition to the waypoints, the mappings from workspace voxels to vertices and edges in the graph, allowing online planning in environments with dynamic obstacles. Instead of the \textit{tensor graph product} in~\cite{Solovey2016}, the authors in~\cite{AndreasVoelz2017} use the \textit{strong graph product} to combine the DRMs, which also considers single-arm motion, i.e. only one of the two robots is moving. This method is tested in simulations with two \textit{KUKA LBR iiwa} and two \textit{Schunk LWA 4P} robots. The roadmap product is computed online, and the planning time increases rapidly with the number of nodes in the single-arm roadmaps. This results in long planning times of about 10\,s already for single-arm roadmaps with 1,000~nodes. It is also possible to plan directly in the high-dimensional space spanned by two robots, i.e. to use a centralized approach. \cite{Shome2019}~uses \textit{RRT*} as a benchmark, which can only compete with \textit{dRRT*} in simple environments with few obstacles. Its success rate is much lower in the test environments where the robot reaches into a bookshelf. In~\cite{Schulman2014} \textit{RRTconnect} is used as a benchmark for \textit{TrajOpt} on an 18-DoF robot. The success rate and computation time are worse than those of \textit{TrajOpt}. For multi-query planning, \cite{Shome2019}~applies \textit{PRM*} on a 15-DoF system, which takes about seven hours for 50,000~nodes. There are also learning-based approaches for dual-arm motion planning. \cite{Salehian2018}~presents an approach using quadratic programming optimization that avoids any collision checking in online computation. Therefor, similar to the potential field approach, an obstacle boundary function formulated in configuration space is learned for self-collision avoidance. They achieve fast online planning times of around 2\,ms, but the approach requires high effort in preprocessing to learn the obstacle boundary function.\newline%
\begin{figure*}%
    \centering%
    \fontsize{8pt}{11pt}\selectfont%
    \subfloat[Real robot.\label{fig:real}]{%
        \includegraphics[height=3.3cm]{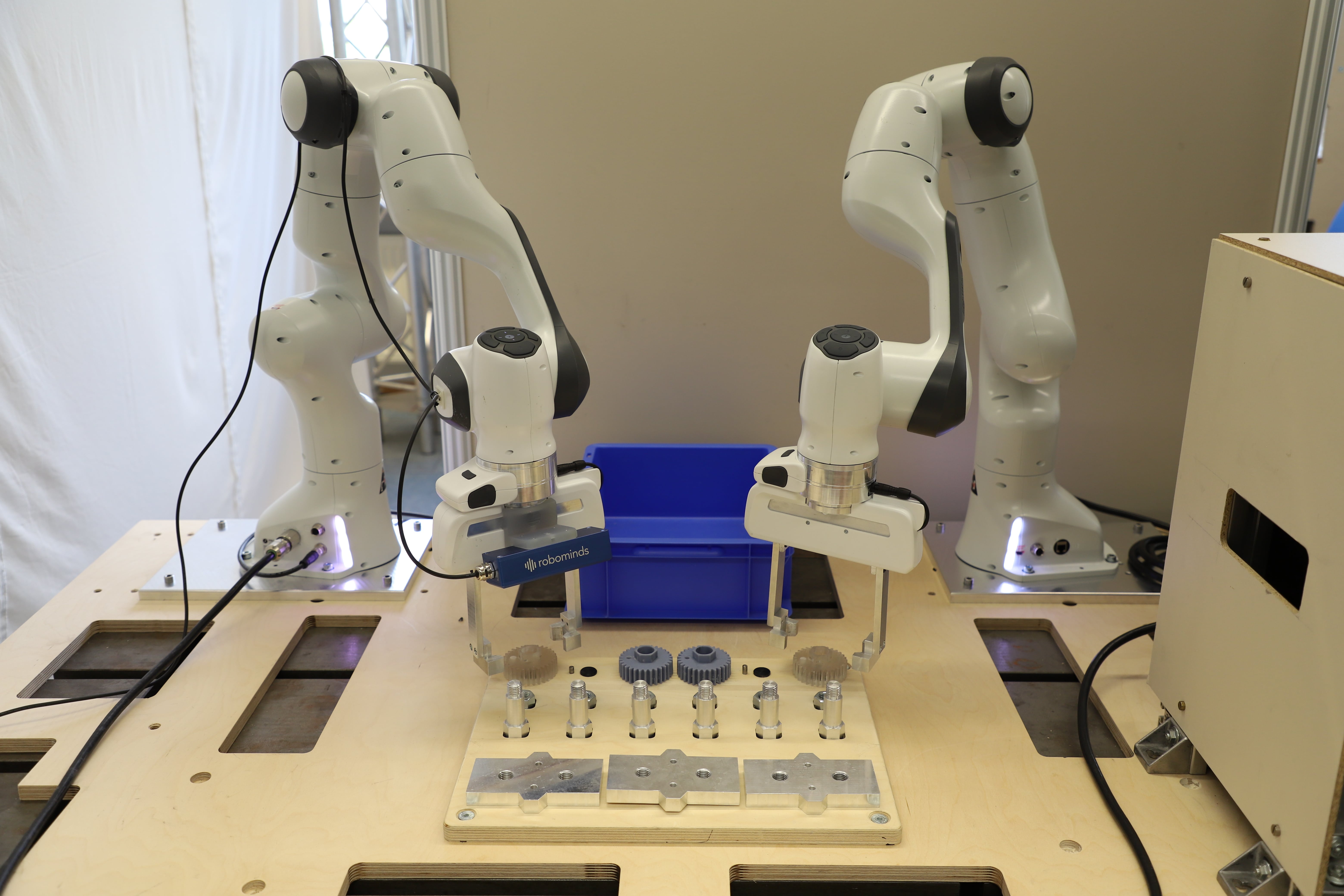}%
	}\hfil%
    \subfloat[\textit{Gazebo} simulation.\label{fig:sim}]{%
        \includegraphics[height=3.3cm]{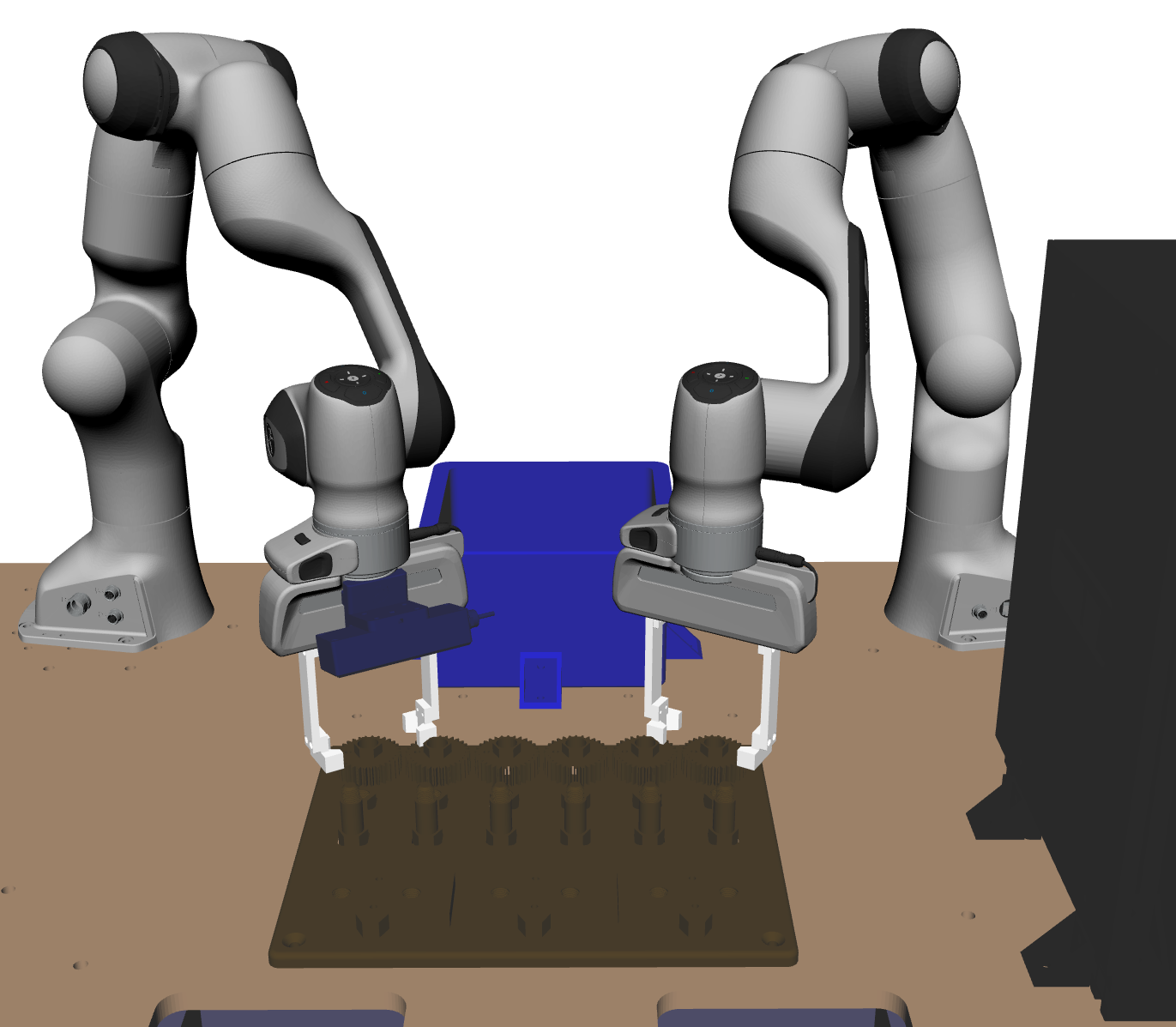}%
	}\hfil%
    \subfloat[\textit{SSV} environment.\label{fig:ssv}]{%
        \includegraphics[height=3.3cm]{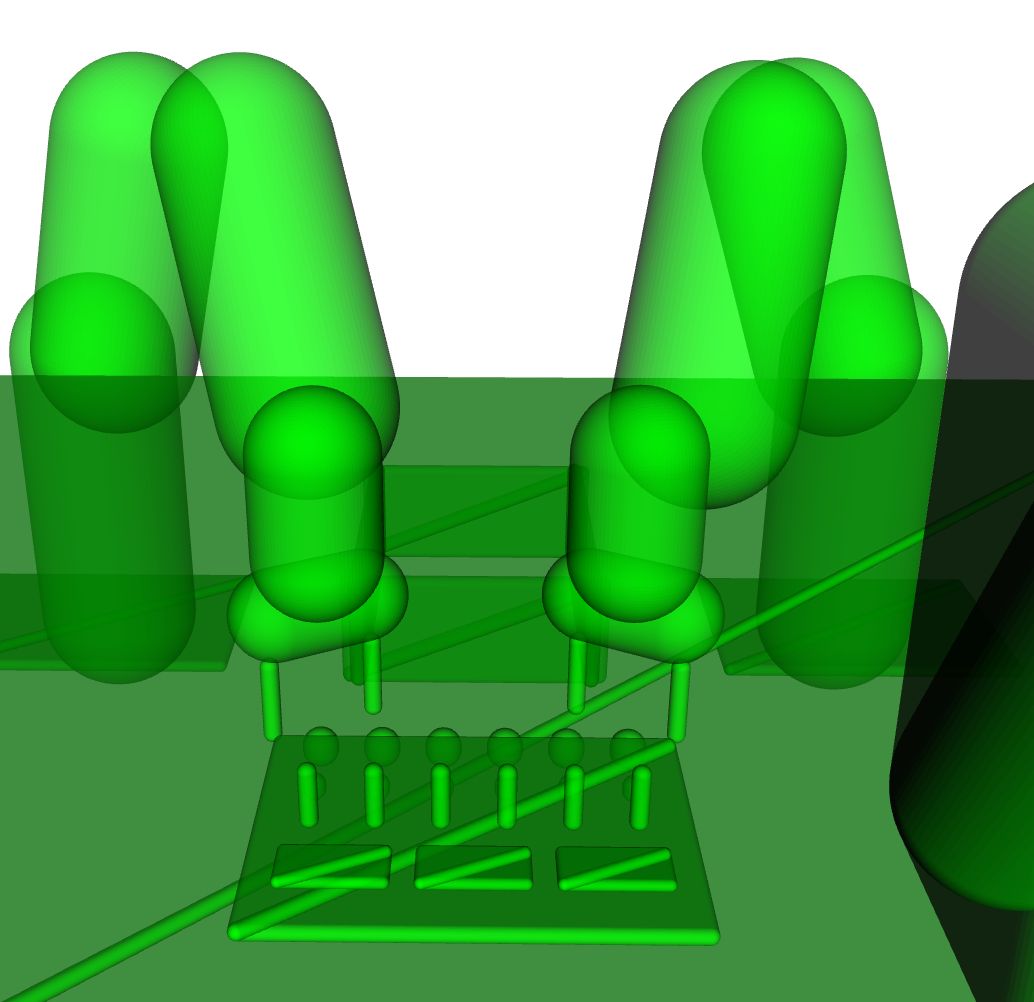}%
	}\hfil%
	\subfloat[\textit{FCL} environment.\label{fig:fcl}]{%
        \includegraphics[height=3.3cm]{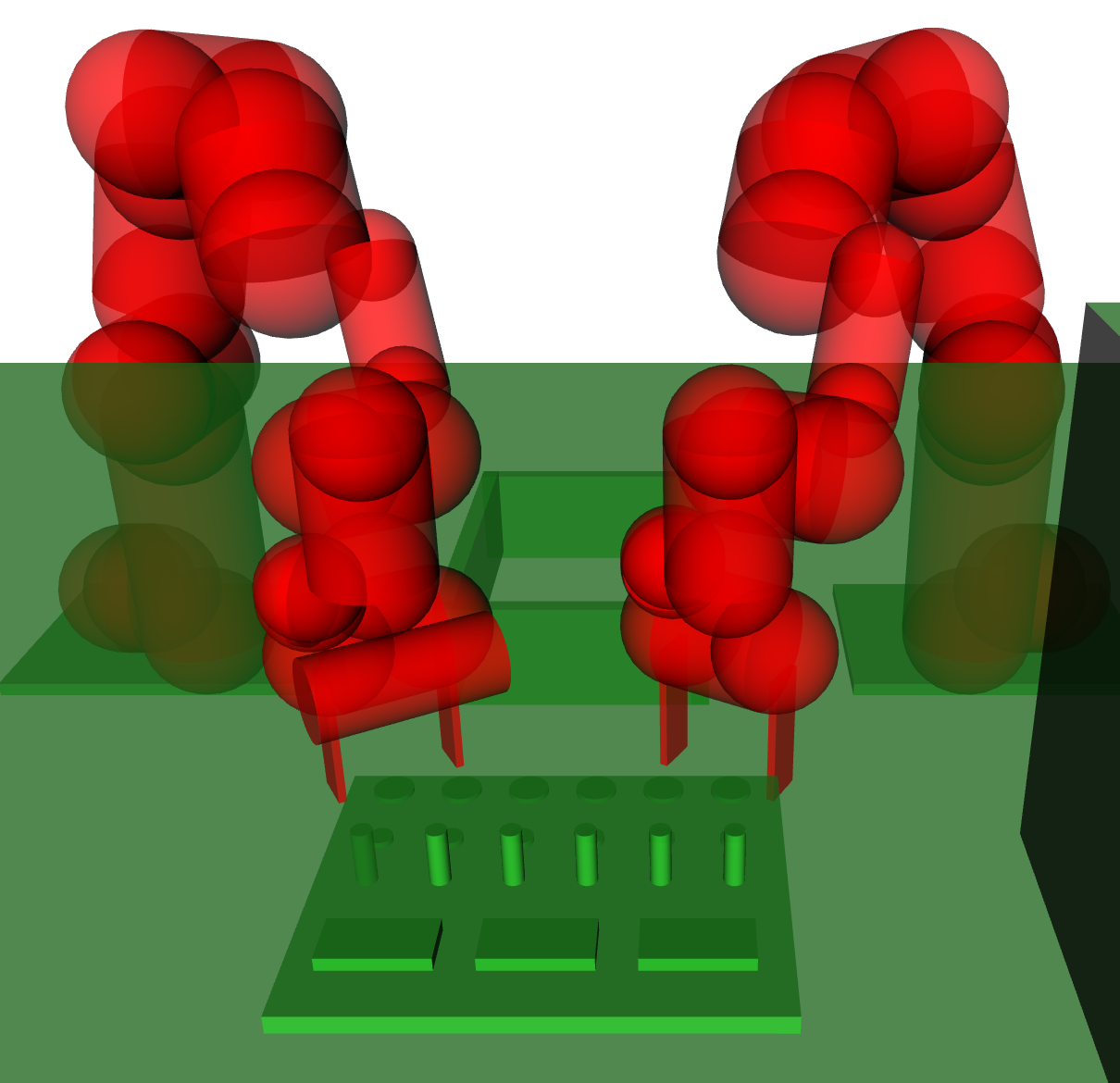}%
	}%
    \caption{Step~\ref{enum:PickAssembly} of the pick-and-place evaluation scenario defined in Section~\ref{sec:PickAndPlace}. We evaluate on the real robot (Figure~\ref{fig:real}) and in simulation (Figure~\ref{fig:sim}) and we use our \textit{SSV}-based environment modeling approach (Figure~\ref{fig:ssv}). The default \textit{FCL}-based environment modeling approach of \textit{MoveIt!} is shown for comparison (Figure~\ref{fig:fcl}).}\label{fig:Scenario}%
\end{figure*}%
Another important aspect of coordinated dual-arm manipulation is task assignment. Which task or pose is assigned to which robot, and when should the robots start executing the task? \cite{Kast2020}~defines three temporal coordination approaches: \emph{sequential} (one robot moves at a time), \emph{parallel} (both robots start their task at the same time), and \emph{asynchronously parallel} (one robot can start a new task while the other robot is still moving). Asynchronous parallelization is implemented in~\cite{Kimmel2017,Kast2020}. \cite{Kimmel2017}~precomputes all tasks and their coordination diagrams offline so that at runtime, only the coordination diagrams are evaluated to move the robots. \cite{Kast2020}~predefines all possible tasks and uses them to determine constraints for an online optimization to determine robot control inputs. Thus, the implementation effort is much higher than for parallel execution. Other approaches avoid idle time by combining all known tasks of a robot arm and computing a parallel execution of these total tasks~\cite{Beuke2018a}. We focus on parallel execution in the following.\newline%
First, most of the presented coordinated planning approaches exclude centralized planning, citing the high computational cost in the augmented configuration space~\cite{Bouvier2022}. Second, most approaches rely on sampling-based planners, which are considered to scale better in multi-robot applications than other approaches~\cite{LaValle2006}. However, few consider an important aspect when using sampling-based approaches: How to avoid jerky paths, i.e. how to improve path quality? Third, an important aspect of real-world applications is often neglected: the continuity of the resulting control commands. We will address these three aspects. Since we operate our robot in a changing environment, we focus on single-query planning and use \textit{RRTconnect}.%
%
%
\section{Benchmark Study}\label{sec:Benchmark}%
This section presents the real-world application scenario for which we developed our proposed motion planning approaches. And it provides information about the third-party software we use to plan the robot motions.%
\subsection{Pick-and-Place Application}\label{sec:PickAndPlace}%
We developed the coordinated planning module for the mobile dual-arm system in Figure~\ref{fig:DualArmSystem}, which we introduce in~\cite{Wittmann2023}. The system is used for transport in intralogistics, and the two arms perform manipulation and pre-assembly operations during transport. The video available at \href{https://youtu.be/tH9bCa7sdIs}{https://youtu.be/tH9bCa7sdIs} shows it in an application scenario defined by an industrial project partner. The main tasks of the two arms are:%
\begin{enumerate}%
    \item Grasp a blue box with both end-effectors at a first workstation.\label{enum:GraspBox}%
    \item Pick up the blue box and place it on the mobile platform.%
    \item Pick up the first five assembly components from the storage on the mobile platform and place them in the blue box.\label{enum:PickAssembly}%
    \item Pick up the last two assembly components and assemble them.%
    \item Grasp the blue box with both end-effectors.\label{enum:GraspBox2}%
    \item Pick up the blue box and place it on a table at a second workstation.%
\end{enumerate}%
Figure~\ref{fig:Scenario} shows the grasping area in the laboratory with its environment modeling. 
\begin{figure*}%
    \centering%
	\def\svgwidth{\textwidth}%
	\resizebox{\textwidth}{!}{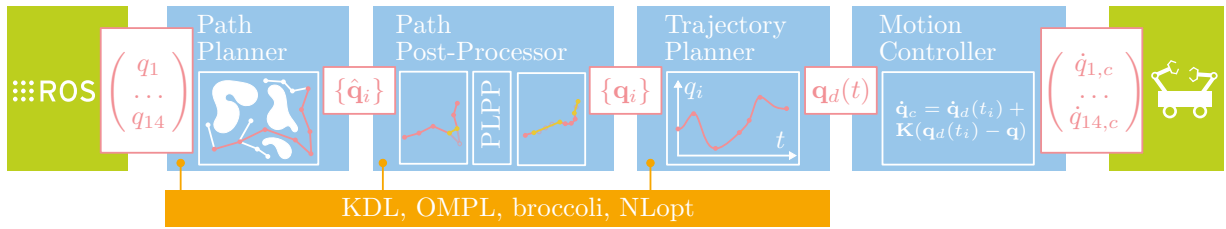}%
    \caption{Planning pipeline of our centralized approach with the four main modules shown in light blue: \textit{Path Planner}, \textit{Path Post-Processor}, \textit{Trajectory Planner}, and \textit{Motion Controller}. The planning pipeline receives motion requests via \textit{ROS} and computes joint velocity commands for the two manipulators.}\label{fig:CentrPipeline}%
\end{figure*}%
This work develops a motion planner for the \textit{coordinated} manipulation of the two arms, i.e. for tasks where both arms manipulate different objects but share the same workspace. This is the case for moving to the grasping poses in steps~\ref{enum:GraspBox} and~\ref{enum:GraspBox2} and for picking up the assembly components in step~\ref{enum:PickAssembly}. 
The other tasks are grasping operations such as closing the grippers with a defined force, or \textit{cooperative} manipulation tasks where both arms manipulate the same object. These steps use a different planning and control approach that we propose in~\cite{Wittmann2023}.\newline%
For the evaluation of this work in Section~\ref{sec:Results}, we will only consider the motion planning for the coordinated motion steps~\ref{enum:GraspBox},~\ref{enum:PickAssembly}, and~\ref{enum:GraspBox2}. In this work, to ensure comparability between different planners, we use hard-coded grasp configurations in the robots' configuration space as goals. In the real application, we would specify the goal pose of the end-effector in Cartesian space and deduce the goal configuration of the robot from it, e.g. using a Newton procedure. Here, we assume that the goal configuration of the robot is already given, i.e. the inverse kinematics on position level has already been computed. This way, we avoid different goal configurations for each motion query, which would result from the position-level inverse kinematics of our redundant 7-DoF manipulators.%
\subsection{Third-Party Software}\label{sec:ThirdParty}%
Our mobile dual-arm system consists of several independent processes that communicate via the \textit{Robot Operating System (ROS)}~\cite{Quigley2009}. For example, one process detects the blue box in the environment using an end-effector-mounted camera, and another process controls the two manipulators. Our motion planning framework for the two manipulators is implemented in C++, and the software architecture is given in~\cite{Wittmann2023}. We use \textit{KDL}~\cite{KDL} to compute the kinematics of the manipulators, and we use \textit{Gazebo}~\cite{Gazebo} to simulate their dynamics. For the \textit{RRTconnect} path planner, we use the implementation of \textit{OMPL}~\cite{OMPL}, and for the spline interpolation of the trajectories, we use \textit{broccoli}~\cite{broccoli}. For distance computations and collision checks, we model the robot and the environment with \textit{Swept Sphere Volumes (SSVs)}. We have implemented this modeling approach in \textit{broccoli} and in~\cite{Wittmann2023}, we show that it is orders of magnitude faster than \textit{FCL}~\cite{FCL}, the default collision checker of \textit{MoveIt!}. Figure~\ref{fig:Scenario} shows both modeling approaches. Our trajectory generation approach and the path post-processing approach that we propose in Section~\ref{sec:PLPP} formulate a constrained nonlinear optimization, and we use \textit{NLopt}~\cite{NLopt2} to solve~it.%
%
%
%
\section{Coordinated Dual-Arm Planning}\label{sec:PlanningMethod}%
This section presents the centralized and the decoupled approaches we benchmark. Their goal is to compute a coordinated collision-free dual-arm manipulation. As Section~\ref{sec:Introduction} mentions, dual-arm planning is currently not supported by \textit{MoveIt!} or other ready-to-use motion planning frameworks, so we implement custom planners.%
\subsection{Centralized Planning}\label{sec:Centralized}%
Since each of the two \textit{Franka Emika} robots has seven DoFs, the centralized approach uses a configuration space of dimension \hbox{2~$\times$~7~$=$~14}. Figure~\ref{fig:CentrPipeline} shows the motion planning pipeline of our centralized approach:%
\begin{figure*}%
    \centering%
	\def\svgwidth{\textwidth}%
	\resizebox{\textwidth}{!}{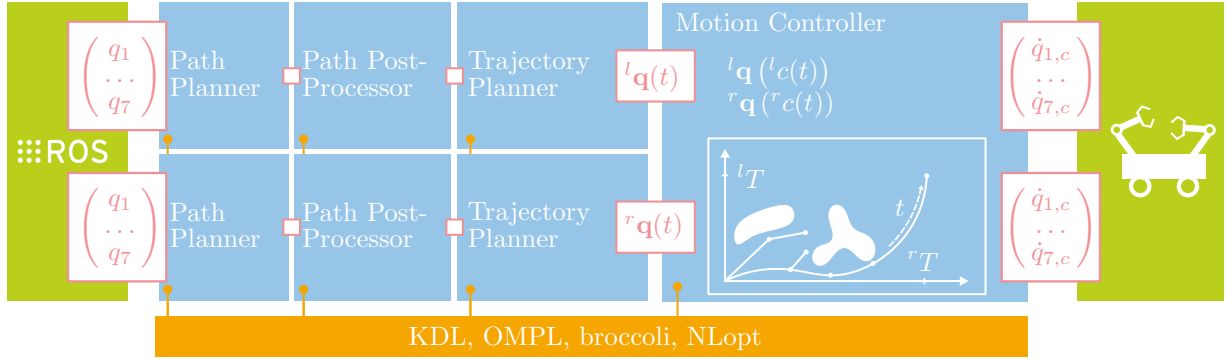}%
    \caption{Planning pipeline of our decoupled approach with the four main modules shown in light blue: \textit{Path Planner}, \textit{Path Post-Processor}, \textit{Trajectory Planner}, and \textit{Motion Controller}. There are two instances of the first three: one for each arm. There is one \textit{Motion Controller} that coordinates the two calculated trajectories. Like the centralized approach in Figure~\ref{fig:CentrPipeline}, the planning pipeline receives motion requests via \textit{ROS} and computes joint velocity commands for the two manipulators.}\label{fig:DecoupledPipeline}%
\end{figure*}%
\begin{enumerate}%
    \item A higher-level task planner or user interface uses \textit{ROS} to communicate desired grasp configurations for both manipulators to the motion planner. The two goal configurations are processed as a 14-DoF vector.%
    \item The centralized \textit{Path Planner} uses \textit{RRTconnect} to find a collision-free path from the current joint configurations to the goal configurations.\label{enum:PathPlanner}%
    \item If a path is found, the path waypoints $\hat{\mathbf{q}}_i$ in the 14-DoF configuration space are passed to the \textit{Path Post-Processor}. This module improves path quality in three steps:%
    \begin{enumerate}
        \item It integrates the default path simplifier of \textit{OMPL}~\cite{SucanSimplify}, which uses approaches such as \textit{Shortcutting} to remove unnecessary waypoints from the solution path~\cite{Geraerts2007}.\label{enum:Simplifier}%
        \item It uses our extended version of the \textit{Path Length Post-Processor (PLPP)}~\cite{Wittmann2022a} presented in Section~\ref{sec:PLPP} to minimize the distance traveled by the end-effector in Cartesian space.%
        \item It applies the \textit{OMPL Path Simplifier} a second time to remove unnecessary waypoints after the optimization of \textit{PLPP}.%
    \end{enumerate}\label{enum:PostProcessor}%
    \item The \textit{Path Post-Processor} passes the modified path waypoints $\mathbf{q}_i$ to the \textit{Trajectory Planner}. We interpolate each DoF with our approach in~\cite{Wittmann2022}, which modifies cubic splines to compute a $\mathcal{C}^2$-continuous trajectory that traverses the waypoints quickly and has zero velocities and zero accelerations at the start and goal.\label{enum:TrajectoryPlanner}%
    \item The computed 14-DoF trajectory $\mathbf{q}_d(t)$ stores an analytical expression of the desired joint configurations, joint velocities, and joint accelerations over time. It is passed to the \textit{Motion Controller}, which evaluates the trajectory at 1\,kHz and which, for a given timestamp $t_i$, computes the joint velocity commands $\dot{\mathbf{q}}_c$ taking into account position-level drift compensation. The joint velocity commands are fed into the hardware interface of the two manipulators.%
\end{enumerate}%
The \textit{Path Planner}, the \textit{Path Post-Processor}, and the \textit{Trajectory Planner} access the third-party functionality described in Section~\ref{sec:ThirdParty} for planning. Applying \textit{PLPP} in step~\ref{enum:PostProcessor} is optional, and in Section~\ref{sec:Results} we will outline the implications.
%
%
\subsection{Decoupled Planning with Fixed-Path Coordination}\label{sec:Decoupled}%
Figure~\ref{fig:DecoupledPipeline} shows the motion planning pipeline of our decoupled approach. As described in Section~\ref{sec:Introduction}, this means that first, a motion is planned separately for each manipulator, ignoring collisions with the other robot, and then the computed motions are coordinated to remove robot-to-robot collisions. Specifically, the planning steps are:%
\begin{enumerate}%
    \item A higher-level task planner or user interface uses \textit{ROS} to communicate the desired grasp configurations for both manipulators to the motion planner. The two goal configurations are processed as two 7-DoF vectors.%
    \item Each of the two grasp configurations is passed to a planning pipeline that is similar to steps~\ref{enum:PathPlanner} to~\ref{enum:TrajectoryPlanner} of the centralized approach in Section~\ref{sec:Centralized}, but that plans in a 7-DoF configuration space instead of a 14-DoF configuration space. Thus, two analytically expressed trajectories $\prescript{l}{}{\mathbf{q}}(t)$ and $\prescript{r}{}{\mathbf{q}}(t)$ are computed, for the left and right manipulator, respectively. We use $l~(left)$ and $r~(right)$ to distinguish both robots, but this is arbitrary, and one could also use $robot~a$/$robot~b$, etc. The planning of the two trajectories uses a multi-threaded approach, i.e. the two planning processes run simultaneously to reduce computation time.%
    \item $\prescript{l}{}{\mathbf{q}}(t)$ and $\prescript{r}{}{\mathbf{q}}(t)$ are passed to the \textit{Motion Controller}, which, upon receiving the two trajectories, computes the fixed-path coordination that prevents robot-to-robot collisions, as described in the remainder of this section.\newline%
    After the fixed-path coordination, similar to the centralized approach in Section~\ref{sec:Centralized}, the \textit{Motion Controller} evaluates the two trajectories at 1\,kHz and forwards the joint velocity commands to both robots.%
\end{enumerate}%
%
The following describes the fixed-path coordination of the trajectories $\prescript{l}{}{\mathbf{q}}(t)$ and $\prescript{r}{}{\mathbf{q}}(t)$ in the \textit{Motion Controller}. The fixed-path coordination is another planning pipeline that uses the same algorithms we use in the \textit{Path Planners}, \textit{Path Post-Processors}, and \textit{Trajectory Planners} of Figures~\ref{fig:CentrPipeline} and~\ref{fig:DecoupledPipeline}:%
\begin{enumerate}%
    \item The space in which the fixed-path coordination plans is the \textit{coordination space}~$\mathcal{C}$, which is two-dimensional for dual-arm manipulation. Let us define for each manipulator a \textit{local} time called~$^l\tau$ and~$^r\tau$, respectively. They refer to pseudo-times that respectively correspond to the trajectories~$^l\mathbf{q}(^l\tau)$ and~$^r\mathbf{q}(^r\tau)$ planned earlier in each of the two separate motion planning pipelines (steps~\ref{enum:PathPlanner} to~\ref{enum:TrajectoryPlanner} in Section~\ref{sec:Centralized}). Considering that these local times can be traversed at different rates,~$\mathcal{C}$ is defined as:%
    \begin{equation}%
        \mathcal{C} = \left\{ \left(\prescript{l}{}{\tau}, \prescript{r}{}{\tau}\right)\mid \prescript{l}{}{\tau}\in[0,\prescript{l}{}{T}], \prescript{r}{}{\tau}\in[0,\prescript{r}{}{T}]\right\} \subset \mathbb{R}^2
    \end{equation}%
    $\prescript{l}{}{\tau}$ and $\prescript{r}{}{\tau}$ are limited by the trajectory duration $\prescript{l}{}{T}$ and $\prescript{r}{}{T}$ that are computed by the \textit{Trajectory Planners}. $\mathcal{C}$ is shown in Figure~\ref{fig:DecoupledPipeline} by the 2D \textit{coordination diagram} in the \textit{Motion Controller}.%
    \item We use the same planning algorithm as for collision-free path planning in the \textit{Path Planners}, i.e. \textit{RRTconnect}, to find collision-free waypoints in $\mathcal{C}$.%
    \item We use the \textit{OMPL Path Simplifier}~\cite{SucanSimplify}, which we also use in the \textit{Path Post-Processors} in Figures~\ref{fig:CentrPipeline}~and~\ref{fig:DecoupledPipeline}, to post-process the waypoints in~$\mathcal{C}$.%
    \item The modified 2D waypoints $\left(\prescript{l}{}{\tau}_i, \prescript{r}{}{\tau}_i\right)$ are interpolated so that we get a continuous mapping of real time $t$ to local trajectory times $\prescript{l}{}{\tau}$ and $\prescript{r}{}{\tau}$:%
    \begin{equation}%
        \left(\begin{array}{c}
            \prescript{l}{}{\tau}\\
        	\prescript{r}{}{\tau}
        \end{array}\right) = \left(\begin{array}{c}
            \prescript{l}{}{c}(t)\\
        	\prescript{r}{}{c}(t)
        \end{array}\right)= \mathbf{c}(t)\label{eq:Coord}
    \end{equation}%
\end{enumerate}%
The fixed-path coordination changes the dynamics of the trajectories computed by the \textit{Trajectory Planners}. Equations~(\ref{eq:c})-(\ref{eq:cddot}) compute the position, velocity, and acceleration profile for each manipulator. They hold for the left and right robot, and we omit the indices for better readability.%
\begin{align}%
    &\mathbf{q}\left(c(t)\right)\label{eq:c}\\%
    &\dot{\mathbf{q}} = \frac{\partial\mathbf{q}\left(c(t)\right)}{\partial c}\cdot\frac{\partial c(t)}{\partial t}\label{eq:cdot}\\%
    &\ddot{\mathbf{q}} = \frac{\partial^2\mathbf{q}\left(c(t)\right)}{\partial c^2}\cdot\left(\frac{\partial c(t)}{\partial t}\right)^2+\frac{\partial \mathbf{q}\left(c(t)\right)}{\partial c}\cdot\frac{\partial ^2c(t)}{\partial t^2}\label{eq:cddot}%
\end{align}%
\begin{figure}%
    \centering%
    \fontsize{8pt}{11pt}\selectfont%
    \subfloat[Position profile in $\mathcal{C}$.\label{fig:c}]{%
		\input{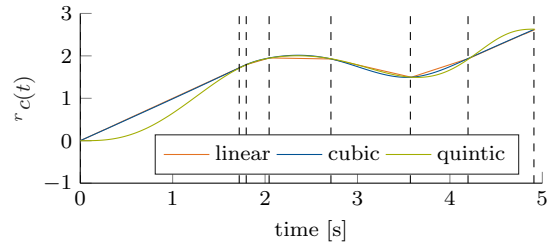}%
	}\\%
    \subfloat[Velocity profile in $\mathcal{C}$.]{%
		\input{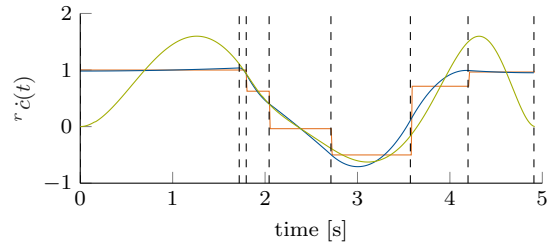}%
	}\\%
    \subfloat[Acceleration profile in $\mathcal{C}$.]{%
		\input{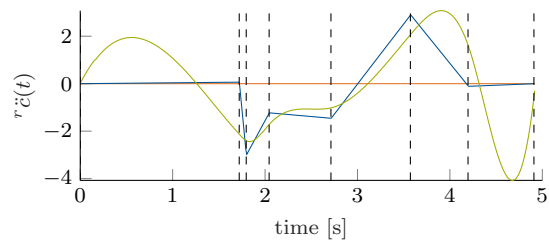}%
	}%
    \caption{Three different interpolation approaches for the coordination space trajectory~$\mathbf{c}(t)$. The black dashed vertical lines show the path waypoints of the fixed-path coordination. That is, these are the coordination time stamps of the right robot computed by the \textit{RRTconnect} and path simplifier in~$\mathcal{C}$.}\label{fig:Coordination}%
\end{figure}%
\subsubsection{Continuous Control Commands}\label{sec:Cont}%
Figure~\ref{fig:Coordination} shows the coordination trajectory profile for the right robot for three different interpolation approaches of the fixed-path coordination: $\mathcal{C}^1$-continuous \textit{linear} interpolation; $\mathcal{C}^2$-continuous \textit{cubic} spline interpolation with accelerations set to zero at the start and goal; $\mathcal{C}^4$-continuous \textit{quintic} spline interpolation with velocities and accelerations set to zero at the start and goal. The separately planned trajectory duration before fixed-path coordination are $\prescript{r}{}{T} = 2.63$\,s and $\prescript{l}{}{T} = 3.61$\,s and the coordinated trajectory duration after fixed-path coordination is $4.91$\,s. The computed path in $\mathcal{C}$ has eight waypoints shown by the black dashed vertical lines (note that there is the start waypoint at 0\,s).\newline%
\begin{figure}%
    \centering%
    \fontsize{8pt}{11pt}\selectfont%
	\input{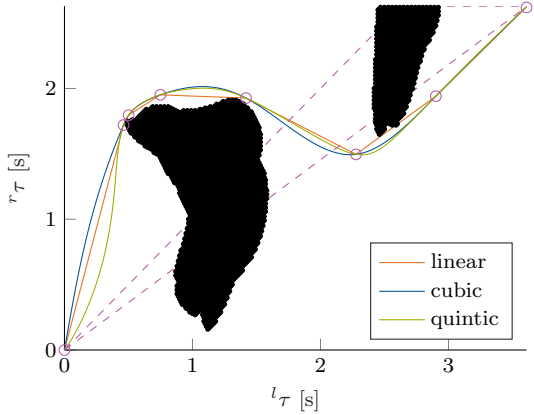}%
    \caption{Coordination diagram for the fixed-path coordination in Figure~\ref{fig:Coordination}. Black areas indicate robot-to-robot collisions in $\mathcal{C}$. The purple circles show the eight waypoints in $\mathcal{C}$. The purple dashed lines correspond to two infeasible solutions that suffer from robot-to-robot collisions: the uncoordinated solution, i.e. without fixed-path coordination, and a solution in which the motion of the right robot is slowed down so that its execution time is the same as that of the left robot, i.e. $3.61$\,s.}\label{fig:CoordDiagr}%
\end{figure}%
Figure~\ref{fig:CoordDiagr} shows the coordination diagram corresponding to Figure~\ref{fig:Coordination}. The orange, blue, and green lines correspond to the three interpolation approaches used in Figure~\ref{fig:Coordination}. The three interpolation approaches avoid collisions, and one can see the difference in continuity between the approaches.\newline%
\begin{figure}%
    \centering%
    \fontsize{8pt}{11pt}\selectfont%
    \subfloat[Position profile.\label{fig:q}]{%
		\input{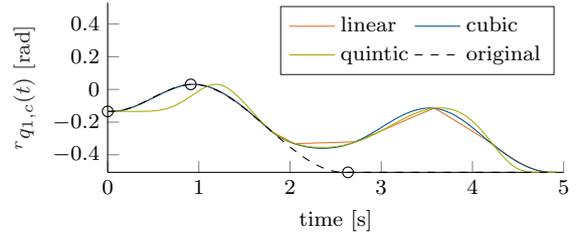}%
	}\\%
    \subfloat[Velocity profile.]{%
		\input{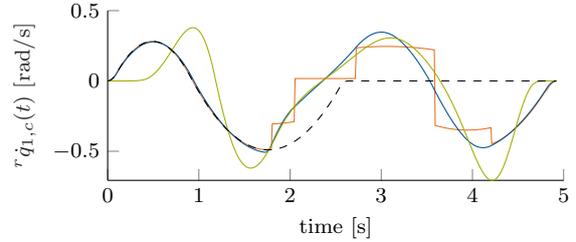}%
	}\\%
    \subfloat[Acceleration profile.]{%
		\input{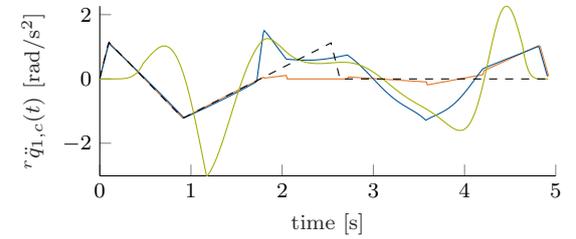}%
	}%
    \caption{The trajectory profiles of the final joint velocity control commands that are computed by the three interpolation methods of the fixed-path coordination in Figures~\ref{fig:Coordination} and~\ref{fig:CoordDiagr}. The orange, blue, and green lines correspond to the three interpolation methods. The black dashed line corresponds to the originally computed trajectory of the first joint before fixed-path coordination. The three black circles in Figure~\ref{fig:q} show the configuration space path waypoints that are computed by \textit{RRTconnect} and modified with the \textit{OMPL Path Simplifier}.}\label{fig:TrajectoryProfile}%
\end{figure}%
Figure~\ref{fig:TrajectoryProfile} shows the resulting trajectory profile of the fixed-path coordination in Figures~\ref{fig:Coordination} and~\ref{fig:CoordDiagr} for the first joint of the right robot. The control trajectories are computed using the approach of Figure~\ref{fig:DecoupledPipeline} and for step~\ref{enum:PickAssembly} of the application described in Section~\ref{sec:PickAndPlace}. This motion is shown in the accompanying video that is available at~\href{https://youtu.be/6vaXqR7Gdew}{https://youtu.be/6vaXqR7Gdew}.\newline%
\cite{Beuke2019,Beuke2018a,ODonnell1989} use linear interpolation in $\mathcal{C}$. But the choice of interpolation should depend on the trajectory generation approach used to compute the two separate uncoordinated trajectories of the two manipulators: The interpolation in $\mathcal{C}$ should not reduce the continuity properties of the trajectories in the robots' configuration space.\newline%
To compute the configuration space trajectories, we use our approach in~\cite{Wittmann2022}. As the dashed black lines in Figure~\ref{fig:TrajectoryProfile} show, it is $\mathcal{C}^2$-continuous, and thus has no discontinuities in the acceleration profile that would lead to oscillations and path tracking inaccuracies. It is also \textit{zero-clamped}, i.e. its desired velocities and accelerations at the first and last waypoints are zero. We want to keep these two characteristics, \textit{smoothness} and \textit{zero-clamped}, also for the final trajectories after the fixed-path coordination.\newline%
Since the approach in~\cite{Wittmann2022} is zero-clamped, $\frac{\partial \mathbf{q}}{\partial c} = \mathbf{0}$ and $\frac{\partial^2\mathbf{q}}{\partial c^2} = \mathbf{0}$ holds in Eqs.~(\ref{eq:cdot}) and~(\ref{eq:cddot}) at the start and goal of the coordinated trajectories. That is, the final motion control commands will be zero-clamped independent of the chosen interpolation order in $\mathcal{C}$. However, the $\mathcal{C}^2$-continuity of the final motion control commands depends on the interpolation order in $\mathcal{C}$ according to the chain rule. The final motion control commands $\mathbf{q}\left(c(t)\right)$ are a composition of two functions, the trajectory interpolation of both manipulators $\mathbf{q}(\tau)$ and the interpolation~$c(t)$ in~$\mathcal{C}$. The former is computed with the approach in~\cite{Wittmann2022} and is $\mathcal{C}^2$-continuous. The composition will only be $\mathcal{C}^2$-continuous if the latter is at least $\mathcal{C}^2$-continuous. Thus, for our case, a \textit{cubic} interpolation in $\mathcal{C}$ is an appropriate choice, which we will use in the remainder of the paper. Figure~\ref{fig:TrajectoryProfile} shows that the \textit{cubic} interpolation gives zero-clamped and $\mathcal{C}^2$-continuous control commands. Higher-order interpolation schemes in~$\mathcal{C}$, such as \textit{quintic}, would also compute smooth and zero-clamped control commands. However, due to the higher order, the profiles oscillate more, and the final control commands are also only $\mathcal{C}^2$-continuous. The \textit{linear} interpolation scheme, as used by~\cite{Beuke2019,Beuke2018a,ODonnell1989}, worsens the continuity properties: The trajectories are $\mathcal{C}^2$-continuous before fixed-path coordination, but they are not after fixed-path coordination, as shown by the discontinuities of the orange line in the velocity profile. As outlined, all trajectories in Figure~\ref{fig:TrajectoryProfile} are zero-clamped regardless of the interpolation order in~$\mathcal{C}$.%
\subsubsection{Deadlocks}\label{sec:Deadlocks}%
\begin{figure}%
    \centering%
	\input{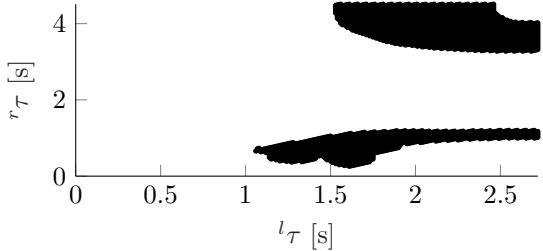}%
    \caption{Coordination diagram for a deadlock situation: There is no collision-free path from start to goal in~$\mathcal{C}$. The durations of the seperately computed trajectories are $\prescript{r}{}{T} = 4.53$\,s and $\prescript{l}{}{T} = 2.72$\,s.}\label{fig:Deadlock}%
\end{figure}%
Fixed-path coordination has one major problem, which is \textit{deadlocks}. These are situations in which \textit{``each manipulator is waiting for the other to proceed"}~\cite{ODonnell1989}. Figure~\ref{fig:Deadlock} shows a coordination diagram of a \textit{deadlock} situation that was generated for the same motion used in Section~\ref{sec:Cont}.\newline%
\cite{ODonnell1989}~proposes a deadlock-free approach by filling non-convexities in the obstacles in~$\mathcal{C}$. This is only applicable if there is a solution in~$\mathcal{C}$, which may not be the case as shown in Figure~\ref{fig:Deadlock}. Furthermore, a possible solution to deadlock in such a situation is to allow the manipulators to \textit{travel back in time}, i.e. to move back to previous configurations along the pre-computed trajectory, which we do. \cite{Beuke2018a} discusses the problem of deadlocks but does not solve it and refers to future work. Deadlocks prevent decoupled planning from achieving \textit{completeness}, so we compare decoupled planning to centralized planning, which does not suffer from this problem.%
%
%
\section{Path Length Minimization in SO(3)}\label{sec:PLPP}%
Both coordination approaches presented in Section~\ref{sec:PlanningMethod} are based on the sampling-based planning method \textit{RRTconnect}. Therefore, the resulting paths are generally jerky with long path lengths. As shown in Figures~\ref{fig:CentrPipeline} and ~\ref{fig:DecoupledPipeline}, we use a \textit{Path Post-Processor} to improve the path quality. Other dual-arm planning frameworks include such approaches already in the path planning phase~\cite{Shome2019,Cohen2014}, but these optimized versions of existing planning algorithms, e.g. \textit{dRRT*}~\cite{Shome2019}, suffer from long computation times because they also optimize in regions of the solution space that are not relevant for the final path. 
Similar to others, e.g. \cite{Shome2019,Geraerts2007,Wittmann2020,Wittmann2022a}, we consider path length as path quality metric because our robots operate next to humans, and therefore, we want to avoid jerky motions, i.e. long, motions. Specifically, we define the distance traveled by the end-effector in Cartesian space as our path quality metric, which we want to minimize. We apply the \textit{OMPL Path Simplifier}~\cite{SucanSimplify} as a first post-processing step. This is a computationally efficient approach based on \textit{Shortcutting}, but its output path still leaves room for improvement. Therefore, we propose using \textit{PLPP} as a second post-processing step.\newline%
In~\cite{Wittmann2022a}, we propose \textit{PLPP}, an algorithm for minimizing path length in Cartesian space. It formulates a nonlinear optimization problem and uses analytic gradients to be computationally efficient. However, we only provided evaluations in a \textit{MATLAB} simulation of a single-arm application. Furthermore, we only proposed the theory for Euclidean spaces, i.e. we only minimize the translational distance of the end-effector motion and the motion in configuration space. The translational part of the Cartesian space and the configuration space are both Euclidean spaces. However, the rotational part of the Cartesian space is not. It is the \textit{special orthogonal group in three dimensions}~$SO(3)$, a nonlinear space. In this work, we extend \textit{PLPP} so that it is also applicable to $SO(3)$, and we integrate it into our C++ framework and provide evaluations on real hardware in a dual-arm scenario.%
\subsection{Path Length Post-Processor}\label{sec:PLPPtrans}%
This subsection repeats the problem formulation of \textit{PLPP} in~\cite{Wittmann2022a} when applied to the end-effector's translational motion in Cartesian space. Equations~(\ref{eq:PLPP_trans_obj}-\ref{eq:PLPP_trans_coll}) formulate the optimization problem.%
\begin{align}%
	\min_{\mathcal{Q}} f(\mathcal{Q}) &= \sum_{i=1}^{n-1}{\frac{1}{2}\left(\mathbf{x}_{i+1}-\mathbf{x}_i\right)^\text{T}\left(\mathbf{x}_{i+1}-\mathbf{x}_i\right)}\span\label{eq:PLPP_trans_obj}\\%
	d(\mathbf{q}_i) &\geq d_{obs}  \quad\forall  i \in [2, n-1]\label{eq:PLPP_trans_coll}%
\end{align}%
$\mathcal{Q}$ denotes the configuration space waypoints of the collision-free path computed by the \textit{Path Planner} in Figures~\ref{fig:CentrPipeline} and~\ref{fig:DecoupledPipeline}. We do not want to modify the first waypoint, the current robot configuration, and also not the last waypoint, the desired goal configuration. Therefore, $\mathcal{Q}$ denotes the $n-2$ intermediate waypoints $\mathbf{q}_i$ that we input to \textit{PLPP}: \hbox{$\mathcal{Q}=\{\mathbf{q}_i \mid\forall~i \in [2, n-1]\}$}. $\mathbf{x}_i = \mathbf{x}(\mathbf{q}_i)$ computes the position $(x_i,y_i,z_i)^\text{T}$ of the end-effector in Cartesian space with forward kinematics. We compute all positions and orientations in Cartesian space w.r.t. a world frame B$_0$, but we omit this additional index in the following for better readability. Equation~(\ref{eq:PLPP_trans_coll}) ensures that each waypoint is at least $d_{obs}$ away from the nearest obstacle, which may be an environment object or a robot link.\newline%
To solve the optimization problem efficiently, we derive the analytic gradients of Eqs.~(\ref{eq:PLPP_trans_obj}-\ref{eq:PLPP_trans_coll}) w.r.t.~$\mathcal{Q}$ in~\cite{Wittmann2022a}. The following holds for the objective function $f$ and $\mathbf{J}^{\text{trans}}_i$ is the translational Jacobian of the end-effector for a given waypoint:%
\begin{equation}%
	\nabla_{\mathcal{Q}}f =[ \dots,~(2\mathbf{x}_i-\mathbf{x}_{i-1}-\mathbf{x}_{i+1})^\text{T}\mathbf{J}^{\text{trans}}_i,~\dots]^\text{T}\label{eq:GradientTrans}%
\end{equation}%
We refer to~\cite{Wittmann2022a} for the derivation and the gradient of the inequality constraint. In~\cite{Wittmann2022a}, \textit{PLPP} was used stand-alone without the \textit{OMPL Path Simplifier}~\cite{SucanSimplify}, so there were additional constraints for better robustness: In the following, in contrast to~\cite{Wittmann2022a}, we do not consider an inequality constraint for joint limits and an inequality constraint for a uniform distribution of the waypoints. Our simulations and experiments have shown that these additional constraints are not necessary, but they can be added, e.g. when the application requires the robot to operate close to its joint limits. Furthermore, in this work, we scale the objective function and the collision inequality constraint. That is, we divide Eq.~(\ref{eq:PLPP_trans_coll}) by $d_{obs}$, and we divide Eq.~(\ref{eq:PLPP_trans_obj}) by the initial Cartesian space path length, i.e. by its value of the first iteration during the optimization.%
\subsection{Derivation of Rotational Distances w.r.t. Quaternions}\label{sec:AnaGradients}%
The problem formulation and the derivation of the analytic gradients in Section~\ref{sec:PLPPtrans} are applicable to Euclidean spaces, i.e. to the configuration space of the robot or to the translational motion of its end-effector in Cartesian space. However, as explained, the formulae do not hold for the rotational motion of the end-effector that takes place in $SO(3)$, a nonlinear space to which a different type of algebra applies.\newline%
This section proposes an additional term for the objective function of \textit{PLPP} in Eq.~(\ref{eq:PLPP_trans_obj}) that minimizes the rotational path length, and it derives the corresponding gradient w.r.t.~$\mathcal{Q}$, similar to Eq.~(\ref{eq:GradientTrans}). We have not found a mathematical presentation of this gradient-based optimization in literature. Most path post-processing approaches are proposed only for configurations space as those discussed in Section~\ref{sec:StateArt}. \cite{Lin2018}~presents a trajectory optimization approach that also considers Cartesian space waypoints, but they ignore rotations. \cite{Geraerts2007}~computes the distance between rotations in~$SO(3)$, but no gradients.\newline%
%
\begin{figure*}%
    \centering%
    \fontsize{8pt}{11pt}\selectfont%
    \subfloat[Minimize the translational waypoint distance only.\label{fig:TransOnly}]{%
        \includegraphics[width=\textwidth]{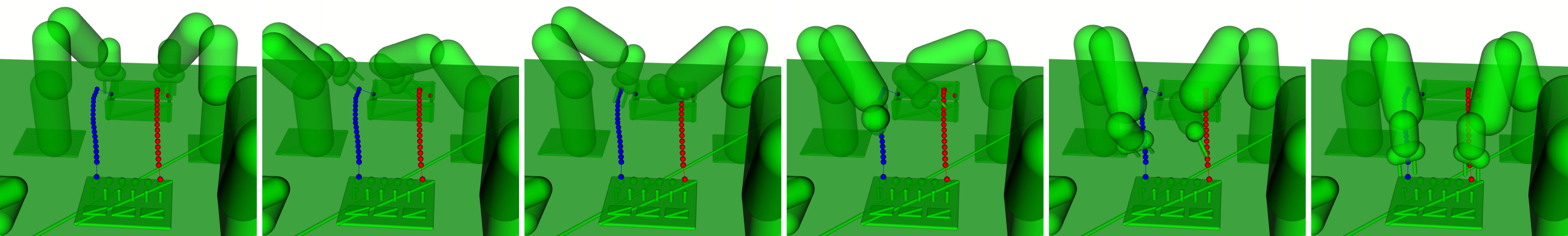}%
	}\\%
    \subfloat[Minimize the translational and rotational waypoint distance.\label{fig:TransAndRot}]{%
        \includegraphics[width=\textwidth]{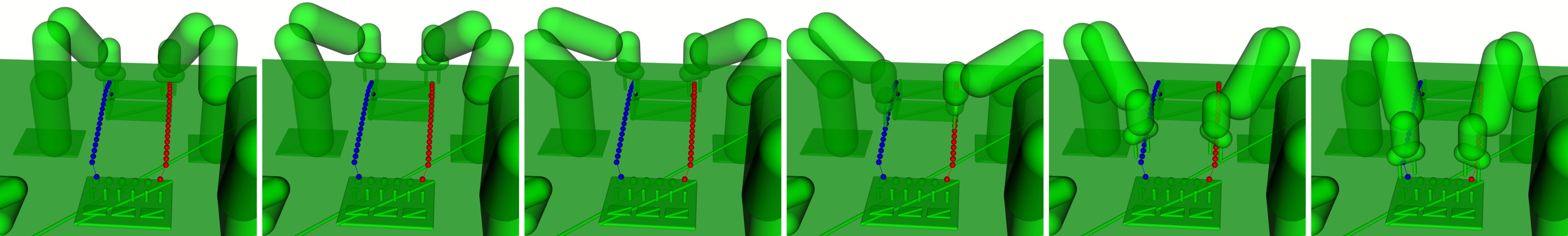}%
	}%
    \caption{Snapshots of a motion between a starting configuration and a goal configuration. Blue spheres are the TCP positions along the optimized path of the left robot; red spheres are those of the right robot. Figure~\ref{fig:TransOnly} shows the approach according to Eq.~(\ref{eq:PLPP_trans_obj}), which ensures a straight connection between the end-effector positions at the start and the goal. However, the rotational motion of the end-effector along the path is large. Figure~\ref{fig:TransAndRot} shows the approach according to Eq.~(\ref{eq:PLPP_new}), which also ensures a straight connection between start and goal but avoids the large rotational motions. For a better visualization, in Figures~\ref{fig:TransOnly} and~\ref{fig:TransAndRot}, we did not apply the \textit{OMPL Path Simplifier} a second time after \textit{PLPP} (see Section~\ref{sec:Centralized}), so there are still the 20~waypoints in the path.\label{fig:Orientation}}%
\end{figure*}%
We define rotational waypoints of the end-effector w.r.t. the world frame B$_0$ and we represent these waypoints with unit quaternions $\prescript{}{0}{\mathbf{u}}_i = [a_i,k_i,l_i,m_i] = [a_i, \mathbf{v}_i^\text{T}]$. $a_i$ and $\mathbf{v}_i$ are the real and imaginary part, respectively. The configuration space waypoints in~$\mathcal{Q}$ are the optimization variables, and the unit quaternions are computed with forward kinematics. Similar to the translational Cartesian distance used for the minimization in Eq.~(\ref{eq:PLPP_trans_obj}), we compute the rotational distance $d^{\text{rot}}_i$ between two adjacent waypoints~$\mathbf{q}_i$ and~$\mathbf{q}_{i+1}$ as follows, where~$\otimes$ denotes the quaternion product:%
\begin{align}%
    \prescript{}{i}{\mathbf{u}}_{i+1} &= \prescript{}{0}{\mathbf{u}}^{-1}_{i}\otimes\prescript{}{0}{\mathbf{u}}_{i+1}\label{eq:rel}\\%
    \prescript{}{i}{\mathbf{p}}_{i+1} &= \text{logMap}(\prescript{}{i}{\mathbf{u}}_{i+1})\label{eq:Log}\\%
    &=\frac{2.0~\text{acos}\left(\operatorname{Re}(\prescript{}{i}{\mathbf{u}}_{i+1})\right)}{\text{sin}\left(\text{acos}\left(\operatorname{Re}(\prescript{}{i}{\mathbf{u}}_{i+1})\right)\right)}\cdot\operatorname{Img}(\prescript{}{i}{\mathbf{u}}_{i+1})\label{eq:Log_impl}\\%
    d^{\text{rot}}_i &= \frac{1}{2}\prescript{}{i}{\mathbf{p}}^\text{T}_{i+1}\prescript{}{i}{\mathbf{p}}_{i+1}\label{eq:PLPP_rot}%
\end{align}%
Equation~(\ref{eq:rel}) computes the rotation between two adjacent waypoints. However, unit quaternions suffer from the antipodal problem, i.e. a unit quaternion and its negative represent the same rotation. So, before applying Eq.~(\ref{eq:rel}), we process~$\prescript{}{0}{\mathbf{u}}_{i+1}$ so that this antipodal problem is solved and~$\prescript{}{i}{\mathbf{u}}_{i+1}$ defines the shortest of the two possible arcs between~$\prescript{}{0}{\mathbf{u}}_{i}$ and~$\prescript{}{0}{\mathbf{u}}_{i+1}$. Equations~(\ref{eq:Log}) and~(\ref{eq:Log_impl}) compute the \textit{logarithmic map} logMap of the rotation~$\prescript{}{i}{\mathbf{u}}_{i+1}$~\cite{Grassia1998}, and we use half of its squared $\text{L}^2$-norm as our distance metric for rotations in Eq.~(\ref{eq:PLPP_rot}). We follow this approach because the $\text{L}^2$-norm of the logarithmic map is intuitive, while the $\text{L}^2$-norm of, for example, the imaginary part of the unit quaternion is not: When the rotation angle around a constant rotation axis doubles, the $\text{L}^2$-norm of the logarithmic doubles, but the $\text{L}^2$-norm of the unit quaternion will not and will instead behave nonlinearly. $\operatorname{Re}(\dots)$ and $\operatorname{Img}(\dots)$ in Eq.~(\ref{eq:Log_impl}) extract the real and imaginary parts of a unit quaternion, respectively.\newline%
To solve the optimization problem efficiently, we must derive the analytic gradient of the total rotational path length w.r.t. all~$\mathbf{q}_i$ in~$\mathcal{Q}$. Thereby, we apply the chain rule to Eq.~(\ref{eq:PLPP_rot}):%
\begin{align}%
    \frac{\partial d^{\text{rot}}_i}{\partial\mathbf{q}_i} &= \frac{\partial\text{(\ref{eq:PLPP_rot})}}{\partial\prescript{}{i}{\mathbf{p}}_{i+1}}\frac{\partial\text{(\ref{eq:Log})}}{\partial\mathbf{q}_i}\label{eq:Chain1}%
\end{align}%
In general, for the variation of logMap in Eq.~(\ref{eq:Log}), the following holds~\cite{Geradin2001}:
\begin{equation}%
    \delta\prescript{}{i}{\mathbf{p}}_{i+1} = \mathbf{T}^{-1}\left(\prescript{}{i}{\mathbf{p}}_{i+1}\right)\delta\boldsymbol{\xi}%
\end{equation}%
$\delta\boldsymbol{\xi}$ are the  three infinitesimal rotations due to a variation of a parameter that~$\delta\prescript{}{i}{\mathbf{p}}_{i+1}$ depends on. For example, when deriving w.r.t. time~$t$, the angular velocity~$\prescript{}{i}{\boldsymbol{\omega}}_{i+1}$ must be used:~\hbox{$\frac{\partial\prescript{}{i}{\mathbf{p}}_{i+1}}{\partial t} = \mathbf{T}^{-1}\left(\prescript{}{i}{\mathbf{p}}_{i+1}\right)\prescript{}{i}{\boldsymbol{\omega}}_{i+1}$}. When deriving w.r.t. the robot joint configuration~$\mathbf{q}$, the rotational part of the Jacobian~$\prescript{}{i}{\mathbf{J}}^{\text{rot}}_{i+1}$ must be used:~\hbox{$\frac{\partial\prescript{}{i}{\mathbf{p}}_{i+1}}{\partial\mathbf{q}} = \mathbf{T}^{-1}\left(\prescript{}{i}{\mathbf{p}}_{i+1}\right)\prescript{}{i}{\mathbf{J}}^{\text{rot}}_{i+1}$}. The matrix $\mathbf{T}^{-1}$ is the Jacobian of the \textit{logarithmic map}, which is the inverse of the Jacobian of the \textit{exponential map}~$\mathbf{T}$. \cite{Geradin2001,Maekinen2008}~provide the latter, but to avoid inversion, we use the analytic expression for~$\mathbf{T}^{-1}$ as derived in~\cite{Chirikjian2012}:%
\begin{align}%
    \mathbf{T}^{-1}(\mathbf{p})&=\mathbf{I}-\frac{1}{2}\Tilde{\mathbf{p}}+\frac{1-\gamma}{\lvert\mathbf{p}\rvert^2}\Tilde{\mathbf{p}}\Tilde{\mathbf{p}}\label{eq:ExpJac}\\%
    \gamma &= \frac{0.5~\lvert\mathbf{p}\rvert}{\text{tan}\left(0.5~\lvert\mathbf{p}\rvert\right)}%
\end{align}%
For the derivation of the gradient of the objective function, we further have to consider that a path waypoint~$\mathbf{q}_i$ is part of its left and right motion segment $d^{\text{rot}}_{i-1}$ and~$d^{\text{rot}}_i$.\newline%
Thus, for the left motion segment~$d^{\text{rot}}_{i-1}$, we obtain:%
\begin{equation}
    \frac{\partial d^{\text{rot}}_{i-1}}{\partial \mathbf{q}_{i}} =\prescript{}{i-1}{\mathbf{p}}^\text{T}_{i}\mathbf{T}^{-1}\left(\prescript{}{i-1}{\mathbf{p}}_{i}\right)\prescript{}{0}{\mathbf{A}}_{i-1}^\text{T}\prescript{}{0}{\mathbf{J}}^{\text{rot}}_{i}\label{eq:left}%
\end{equation}%
$\prescript{}{0}{\mathbf{A}}_{i-1}$ is the rotation matrix that describes the orientation of the end-effector at waypoint~$\mathbf{q}_{i-1}$ w.r.t.~B$_0$. Similar, for the right motion segment~$d^{\text{rot}}_i$, we obtain:%
\begin{align}%
    \frac{\partial d^{\text{rot}}_i}{\partial\mathbf{q}_i} &= \prescript{}{i}{\mathbf{p}}^\text{T}_{i+1}\frac{\partial\prescript{}{i}{\mathbf{p}}_{i+1}}{\partial\mathbf{q}_i}=\prescript{}{i}{\mathbf{p}}^\text{T}_{i+1}\left(-\frac{\partial\prescript{}{i+1}{\mathbf{p}}_{i}}{\partial\mathbf{q}_i}\right)\nonumber\\%
    &= -\prescript{}{i}{\mathbf{p}}_{i+1}^\text{T}\mathbf{T}^{-1}\left(-\prescript{}{i}{\mathbf{p}}_{i+1}\right)\prescript{}{0}{\mathbf{A}}_{i+1}^\text{T}\prescript{}{0}{\mathbf{J}}^{\text{rot}}_i\label{eq:right}%
\end{align}%
The analytical Jacobians of the \textit{logarithmic map} and \textit{exponential map} are given in~\cite{Chirikjian2012}. Still, we have not found a formulation to compute gradient information using unit quaternions according to Eqs.~(\ref{eq:Log}),~(\ref{eq:ExpJac}),~(\ref{eq:left}), and (\ref{eq:right}) in literature.\newline\newline%
With \textit{PLPP}, we want to minimize the Cartesian path length while keeping a defined distance to obstacles, so the final objective function is:%
\begin{equation}%
	\frac{1}{2}\sum_{i=1}^{n-1}{\left(\alpha\cdot\left(\mathbf{x}_{i+1}-\mathbf{x}_i\right)^\text{T}\left(\mathbf{x}_{i+1}-\mathbf{x}_i\right) + \prescript{}{i}{\mathbf{p}}^\text{T}_{i+1}\prescript{}{i}{\mathbf{p}}_{i+1}\right)}\label{eq:PLPP_new}%
\end{equation}%
$\alpha$ balances the different magnitudes of translation and rotation, and we set $\alpha = 5.0$. However, $\alpha$~can be adjusted to suit the given application, e.g. to focus more on translations or rotations.\newline%
In gradient-based optimization, it is recommended to ensure the correct implementation of the user-provided gradients since incorrect gradients severely impact the solution algorithm~\cite{NLopt}. We compared our analytically derived gradients to a forward finite difference approximation with a step size of $1\mathrm{e}{-8}\text{ s}$: A relative error of less than $1\mathrm{e}{-6}$ was found, showing that our formulation and implementation of the gradients is correct.\newline%
One might argue that it is sufficient to minimize only the translational distances, i.e. to use Eq.~(\ref{eq:PLPP_trans_obj}) instead of Eq.~(\ref{eq:PLPP_new}). Figure~\ref{fig:Orientation} shows the benefit of including rotations. Ignoring the additional term leads to large rotational motions of the end-effector that are less intuitive for human coworkers. The corresponding motions to the snapshots in Figure~\ref{fig:Orientation} are also provided in the accompanying video at \href{https://youtu.be/6vaXqR7Gdew}{https://youtu.be/6vaXqR7Gdew}. The video further shows that the large rotational motions that must be interpolated also result in a longer motion duration than the approach that considers rotations.%
%
%
\section{Results}\label{sec:Results}%
This section evaluates the centralized and decoupled approaches of Section~\ref{sec:PlanningMethod} and the post-processing approach \textit{PLPP} of Section~\ref{sec:PLPP} in simulation and experiment.%
\subsection{Planner Parameters}%
The computation times of motion queries are very sensitive to the chosen parameterization of the planning pipeline. The following parameters are those that we consider most critical in terms of planning success and computation time:%
\begin{itemize}
    \item We use the \textit{relative function tolerance} $\epsilon_\text{rel}~=~0.001$ as a stopping criterion for the \textit{SQP} solver of \textit{PLPP}. That is, the path optimization will stop when the current objective function value $L$ changes by less than $\epsilon_\text{rel} \cdot L$ in the next iteration. Reducing $\epsilon_\text{}$ would result in slightly better path quality but longer computation times.%
    \item The \textit{RRTconnect} algorithm in the \textit{Path Planner} uses a local planner to verify that a connection between two sampled waypoints is collision-free. This local planner performs a discrete validation of the PTP connection between the two waypoints, and the user must specify the discretization size, i.e. the \textit{longest valid segment fraction}. This is a fraction of the maximum extent of the planning space, and this extent is defined by the sum of the maximum range of motion of each joint. We set this fraction to~$0.001$ for the decoupled approach. The maximum extent of the planning space in the centralized approach is twice that of the decoupled approach. Therefore, to ensure the same longest valid segment length for both approaches, we use half of the \textit{longest valid segment fraction} for the centralized approach. Reducing the \textit{longest valid segment fraction} would result in fewer planning failures but longer computation times.%
    \item Before running \textit{PLPP}, we use the default path simplifier of \textit{OMPL} that runs different computationally cheap shortcutting algorithms. Not doing so results in more jerky paths with more waypoints and thus longer computation times for the subsequent \textit{PLPP}. We also use the \textit{OMPL} simplifier after running the \textit{PLPP} to remove unnecessary segments in the optimized path.%
    \item We interpolate the input path of \textit{PLPP} so that it has a minimum number of 20 waypoints. This way, we maintain a somewhat dense distribution of the waypoints in Cartesian space during the optimization, avoiding collisions due to waypoints that are too far apart.%
\end{itemize}%
\begin{table*}[ht]%
    \small%
    \centering%
    \caption{Centralized vs. decoupled approach without \textit{PLPP} for 800 motion queries.}\label{table:WithoutPLPP}%
    \begin{tabular}{@{}r|l|c|cc@{}}%
        \toprule%
        \multicolumn{1}{r}{}&\multicolumn{1}{c}{}&\multicolumn{1}{|c|}{Centralized}  & \multicolumn{2}{c}{Decoupled}\\%
        \midrule%
        1&Robot ID                                      &                       & \textit{left}     & \textit{right}\\%
        \midrule%
        2&Path Planner Time [ms]                        & 90 $\pm$ 106          & 23 $\pm$ 21       & 14 $\pm$ 9\\%
        3&OMPL Path Simplifier Time [ms]                & 168 $\pm$ 59          & 51 $\pm$ 25       & 50 $\pm$ 28\\%
        4&Trajectory Planner Time [ms]                  & 2.86 $\pm$ 2.36       & 0.75 $\pm$ 0.71   & 0.61 $\pm$ 0.63\\%
        5&Coordination Time [ms]                        & -                     & \multicolumn{2}{c}{373 $\pm$ 90}\\%
        \midrule%
        6&Motion Planning Time [s]                      & 0.26 $\pm$ 0.14       & \multicolumn{2}{c}{0.46 $\pm$ 0.10}\\%
        7&Distance Computation Time [s]                 & 0.25 $\pm$ 0.14       & \multicolumn{2}{c}{0.43 $\pm$ 0.10}\\%
        8&Motion Duration [s]                           & 3.11 $\pm$ 0.93       & \multicolumn{2}{c}{2.28 $\pm$ 1.15}\\%
        \midrule%
        9&Collision Failure                             & 138                    & \multicolumn{2}{c}{103}\\%
        10&Coordination Failure                         & -                     & \multicolumn{2}{c}{232}\\%
        \botrule%
    \end{tabular}%
\end{table*}%
\subsection{Evaluation in Simulation}%
This section evaluates the approaches in simulation using the scenario presented in Section~\ref{sec:Benchmark}. We run the computations on a 64-bit Ubuntu 20.04 machine with 32~GB~RAM and with a 12-core Intel~i7-7800K~CPU running at 3.7~GHz.%
\subsubsection{Without PLPP}\label{sec:Without}%
First, we benchmark the centralized and decoupled approaches without using \textit{PLPP}. We use the scenario of Section~\ref{sec:Benchmark}, i.e. we compute and simulate the motions for steps~\ref{enum:GraspBox},~\ref{enum:PickAssembly}, and~\ref{enum:GraspBox2}. Thus, one cycle is defined by eight different coordinated dual-arm manipulation motion queries:%
\begin{enumerate}%
    \item Grasping poses to pick up the blue box from the table.%
    \item Grasping poses to pick up one gear with each robot.%
    \item Drop-off poses above the blue box.%
    \item Grasping poses to pick up one bolt with each robot.\label{enum:bolts}%
    \item Drop-off poses above the blue box.%
    \item Grasping poses to pick up one shaft with each robot.%
    \item Drop-off poses above the blue box.%
    \item Grasping poses to pick up the blue box from the mobile platform.%
\end{enumerate}%
We ran this cycle 100~times for the centralized and decoupled approaches, i.e. we ran 800 motion queries each, and Table~\ref{table:WithoutPLPP} shows the results. Lines~2-5 give the means and standard deviations of the computation times of the four main modules in the motion planning pipeline according to Figures~\ref{fig:CentrPipeline} and~\ref{fig:DecoupledPipeline}. For the decoupled approach, line~1 distinguishes the computation times for the left and right robots. However, the planning of these two trajectories runs in a multi-threaded approach, as described in Section~\ref{sec:Decoupled}, so the total motion planning time is not the sum of the two approaches but the maximum of both. Lines~6-8 give the total motion planning time, the time spent on distance computations due to collision checks, and the total duration of the computed motion. Lines~9 and~10 give the number of failures among the 800 motion queries.\newline%
We run 800 motion queries composed of the eight different queries listed above. We also use sampling-based motion planning, which computes different solutions for the same motion request in each run due to the randomized approach. For these two reasons, averaging over the 800 queries results in relatively high standard deviations. However, we also looked at the results for one specific motion query out of the eight, and the conclusions and qualitative results that follow are the same, but with a smaller standard deviation.\newline%
The \textit{Motion Planning Time} in line~6 is slightly larger than the sum of the \textit{Path Planner Time}, \textit{OMPL Path Simplifier Time}, \textit{Trajectory Planner Time}, (and \textit{Coordination Time},) because the motion planning pipeline includes other operations such as conversion between different path formats, logging, visualization, etc. For the centralized and decoupled approaches, the \textit{Motion Planning Times} are an order of magnitude smaller than the \textit{Motion Duration Times}, showing that our implemented motion planning pipeline is applicable to online motion planning. In online motion planning, the motion planning time should be in the order of magnitude of the motion execution time~\cite{Wittmann2022a}. For both approaches, the dominant factor in the \textit{Motion Planning Time} is the \textit{Distance Computation Time}, which accounts for more than 90\,\% of the planning time, which is expected for sampling-based motion planning~\cite{Elbanhawi2014,Wittmann2022a}.\newline%
As expected, the computation times for path planning, path post-processing, and trajectory planning are smaller for the decoupled approach because the search space dimension is only half the size. This is not a linear relationship, e.g. for path planning there is a factor of about five between the centralized and decoupled approaches. This is due to the exponential scaling of the computation time w.r.t. the number of DoFs mentioned in Section~\ref{sec:Introduction}.\newline%
\begin{table*}[ht]%
    \small%
    \centering%
    \caption{Centralized vs. decoupled approach with \textit{PLPP} for 800 motion queries.}\label{table:WithPLPP}%
    \begin{tabular}{@{}r|l|c|cc@{}}%
        \toprule%
        \multicolumn{1}{r}{}&\multicolumn{1}{c}{}&\multicolumn{1}{|c|}{Centralized}         & \multicolumn{2}{c}{Decoupled}\\%
        \midrule%
        1&Robot ID                                                  &                       & \textit{left}     & \textit{right}\\%
        \midrule%
        2&Path Planner Time [ms]                                    & 102 $\pm$ 126         & 23 $\pm$ 20       & 15 $\pm$ 9\\%
        3&OMPL Path Simplifier Time [ms]                            & 340 $\pm$ 90          & 104 $\pm$ 33      & 110 $\pm$ 36\\%
        4&\textbf{PLPP Time [ms]}                                   & 1955 $\pm$ 1040       & 318 $\pm$ 218     & 240 $\pm$ 177\\%
        5&Trajectory Planner Time [ms]                              & 5.69 $\pm$ 4.39       & 1.76 $\pm$ 1.12   & 1.78 $\pm$ 1.15\\%
        6&Coordination Time [ms]                                    & -                     & \multicolumn{2}{c}{464 $\pm$ 125}\\%
        \midrule%
        7&Motion Planning Time [s]                                  & 2.73 $\pm$ 2.15       & \multicolumn{2}{c}{0.94 $\pm$ 0.28}\\%
        8&Distance Computation Time [s]                             & 0.57 $\pm$ 0.23       & \multicolumn{2}{c}{0.68 $\pm$ 0.16}\\%
        9&Motion Duration [s]                                       & 2.46 $\pm$ 0.73       & \multicolumn{2}{c}{2.27 $\pm$ 0.48}\\%
        \midrule%
        10&\textbf{Original Path Length (\ref{eq:PLPP_new})}        & 1.80 $\pm$ 1.29       & \multicolumn{2}{c}{1.08 $\pm$ 0.77}\\%
        11&\textbf{Modified Path Length (\ref{eq:PLPP_new})}        & 0.41 $\pm$ 0.19       & \multicolumn{2}{c}{0.49 $\pm$ 0.26}\\%
        \midrule%
        12&Collision Failure                                        & 52                    & \multicolumn{2}{c}{24}\\%
        13&Coordination Failure                                     & -                     & \multicolumn{2}{c}{52}\\%
        \botrule%
    \end{tabular}%
\end{table*}%
As discussed in Section~\ref{sec:Introduction}, decoupled approaches are preferred over centralized approaches in literature because they are considered more computationally efficient, as they scale better with the number of DoFs. However, our research shows this is not the case for dual-arm planning. One must also consider the additional time required to coordinate the separately computed trajectories in the decoupled approach. And in our scenario, the \textit{Coordination Time} makes the decoupled approach (0.46\,s \textit{Motion Planning Time}) more computationally expensive than the centralized approach (0.26\,s \textit{Motion Planning Time}).\newline%
As line~9 shows, both planning approaches suffer from \textit{Collision Failures}, i.e. infeasible paths. This is due to the trajectory interpolation: As explained above, the \textit{Path Planner} uses a local planner that checks for feasibility on a PTP connection between two states. This standard approach in collision-free path planning ensures fast computation times. However, the subsequent \textit{Trajectory Planner} uses a cubic spline interpolation, as explained above. Depending on the trajectory dynamics and the waypoint distances, the cubic interpolation may intersect with obstacles not detected by the PTP feasibility check. For the trajectory dynamics, we use realistic values, i.e. the maximum joint velocity and acceleration of the \textit{MoveIt!} settings for the \textit{Franka Emika} robot. The high number of \textit{Collision Failures} is due to the challenging scenario with thin walls and small fingers moving into the box and around the storage with the gearbox components (see Figure~\ref{fig:ssv}).\newline%
The main drawback of the decoupled approach outlined in Section~\ref{sec:Deadlocks} becomes apparent: This standard motion planning pipeline (\textit{RRTconnect}, \textit{OMPL Path Simplifier}, \textit{Trajectory Interpolation}), which is also used by default in \textit{MoveIt!} (it uses a slightly different trajectory interpolation), leads to a high number of \textit{Coordination Failures}, making the decoupled approach inapplicable. This high number is due to long motion segments around the grasping points so that both end-effectors block each other.\newline%
Neither approach suffers from failure types other than the two listed in Table~\ref{table:WithoutPLPP}, such as non-converging optimization, etc., demonstrating the robustness of our implementation.\newline%
Due to the higher computation time, the lower robustness, and the higher implementation effort of the decoupled approach, we suggest using the centralized approach instead for dual-arm manipulation. For many robots, there will be a break-even point where the motion planning time of the decoupled approach becomes smaller than that of the centralized approach. However, dual-arm systems are the most dominant multi-arm manipulation systems, and the coordination problem remains.%
\subsubsection{With PLPP}\label{sec:With}%
\begin{figure*}%
    \centering%
    \fontsize{8pt}{11pt}\selectfont%
    \subfloat[\textit{RRTconnect} outputs a collision-free path with five waypoints.]{%
        \includegraphics[width=0.19\textwidth]{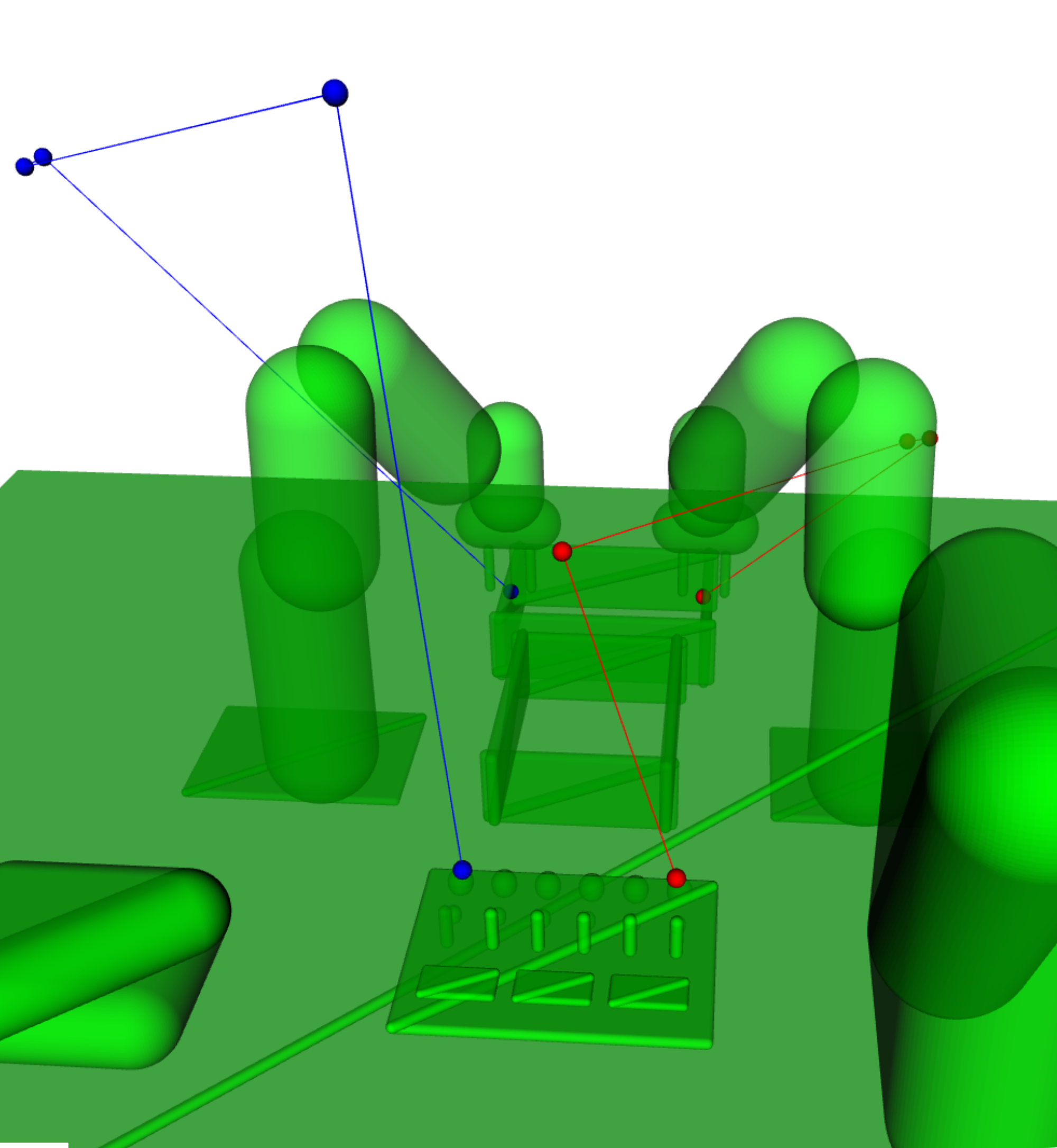}%
	}\hfil%
    \subfloat[The \textit{OMPL Path Simplifier} shortcuts the path to four waypoints.\label{fig:OMPLShorcut}]{%
        \includegraphics[width=0.19\textwidth]{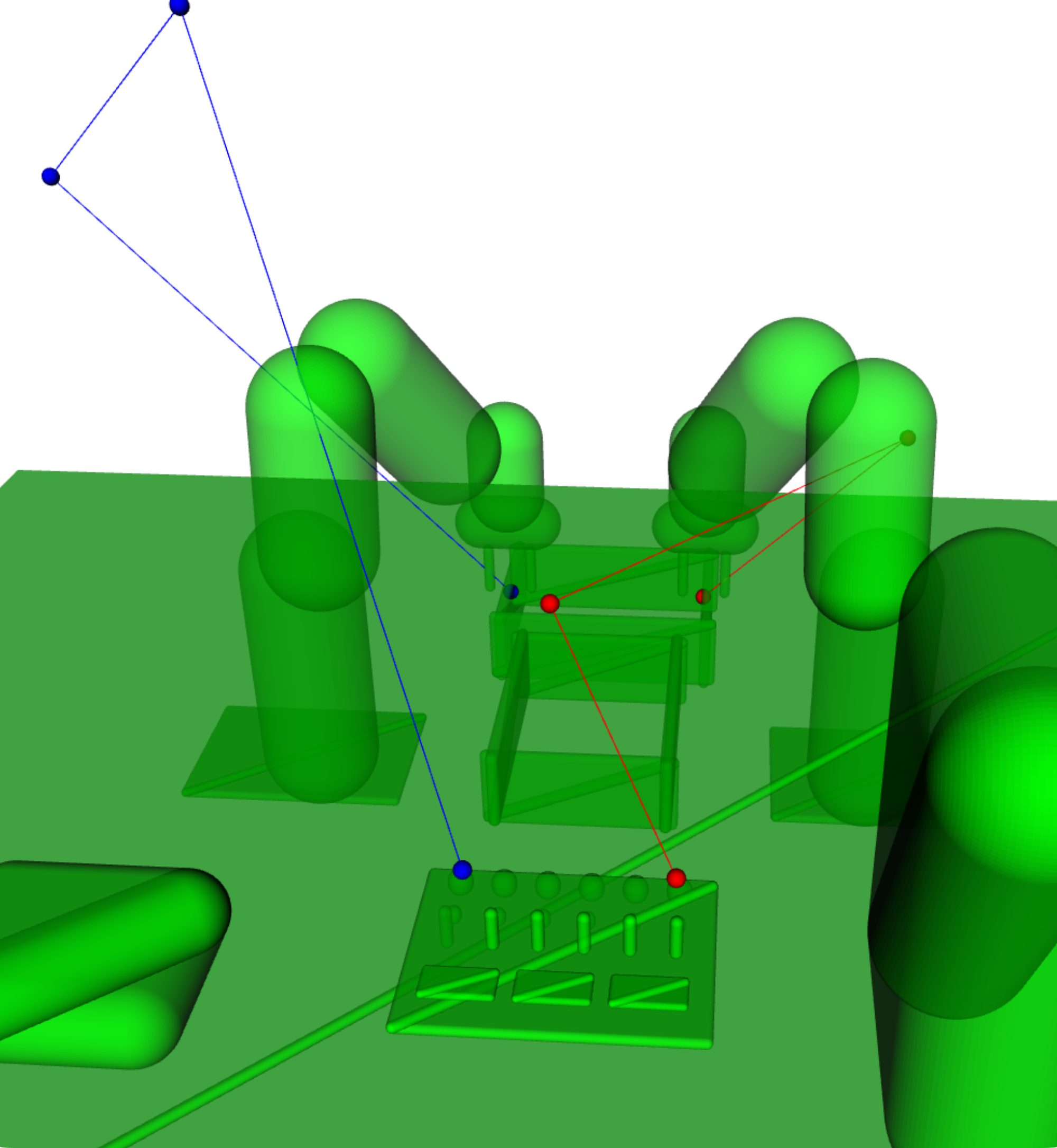}%
	}\hfil%
    \subfloat[We interpolate the path so that it has 20 waypoints.]{%
        \includegraphics[width=0.19\textwidth]{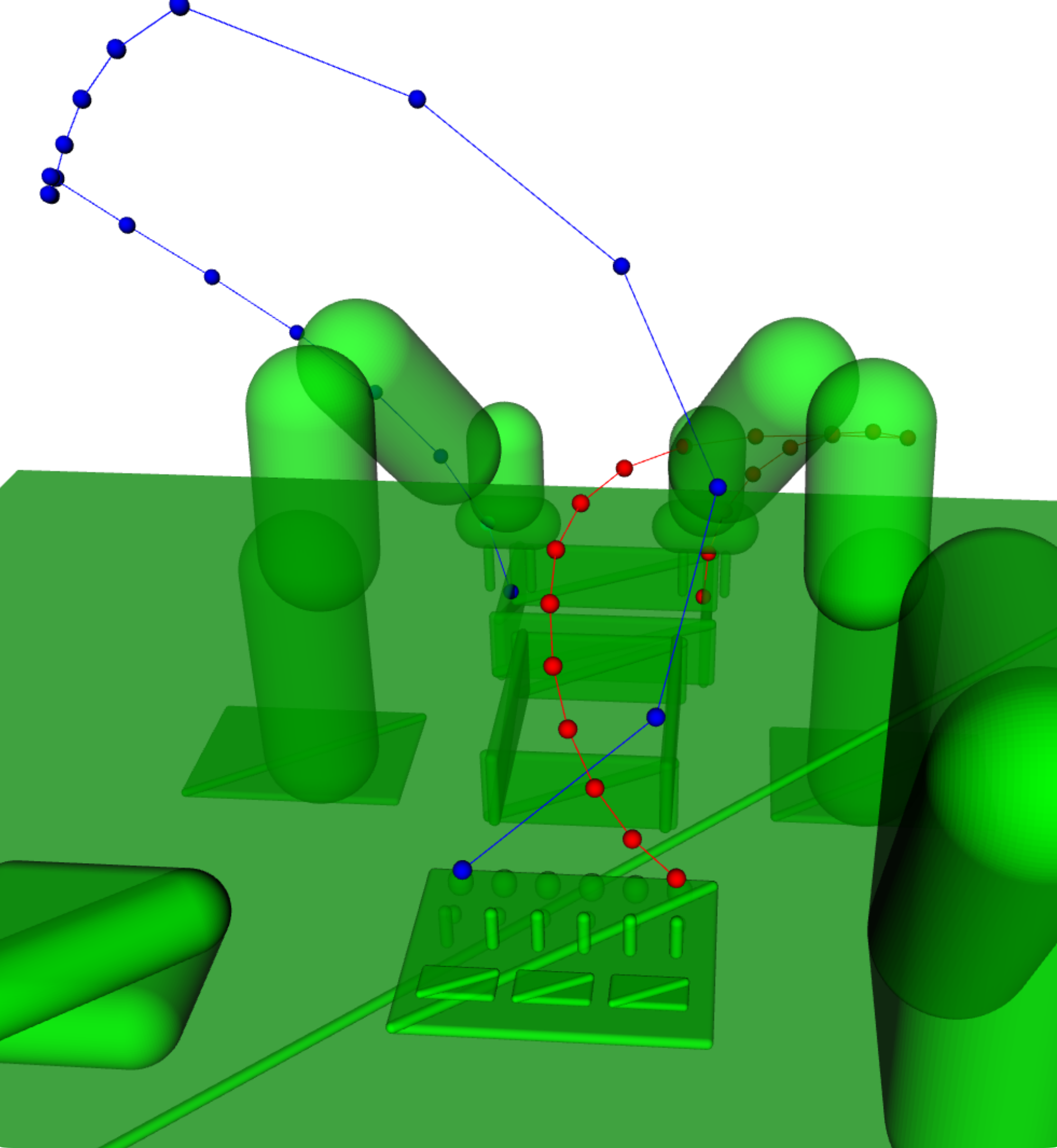}%
	}\hfil%
    \subfloat[\textit{PLPP} minimizes the Cartesian space distance.]{%
        \includegraphics[width=0.19\textwidth]{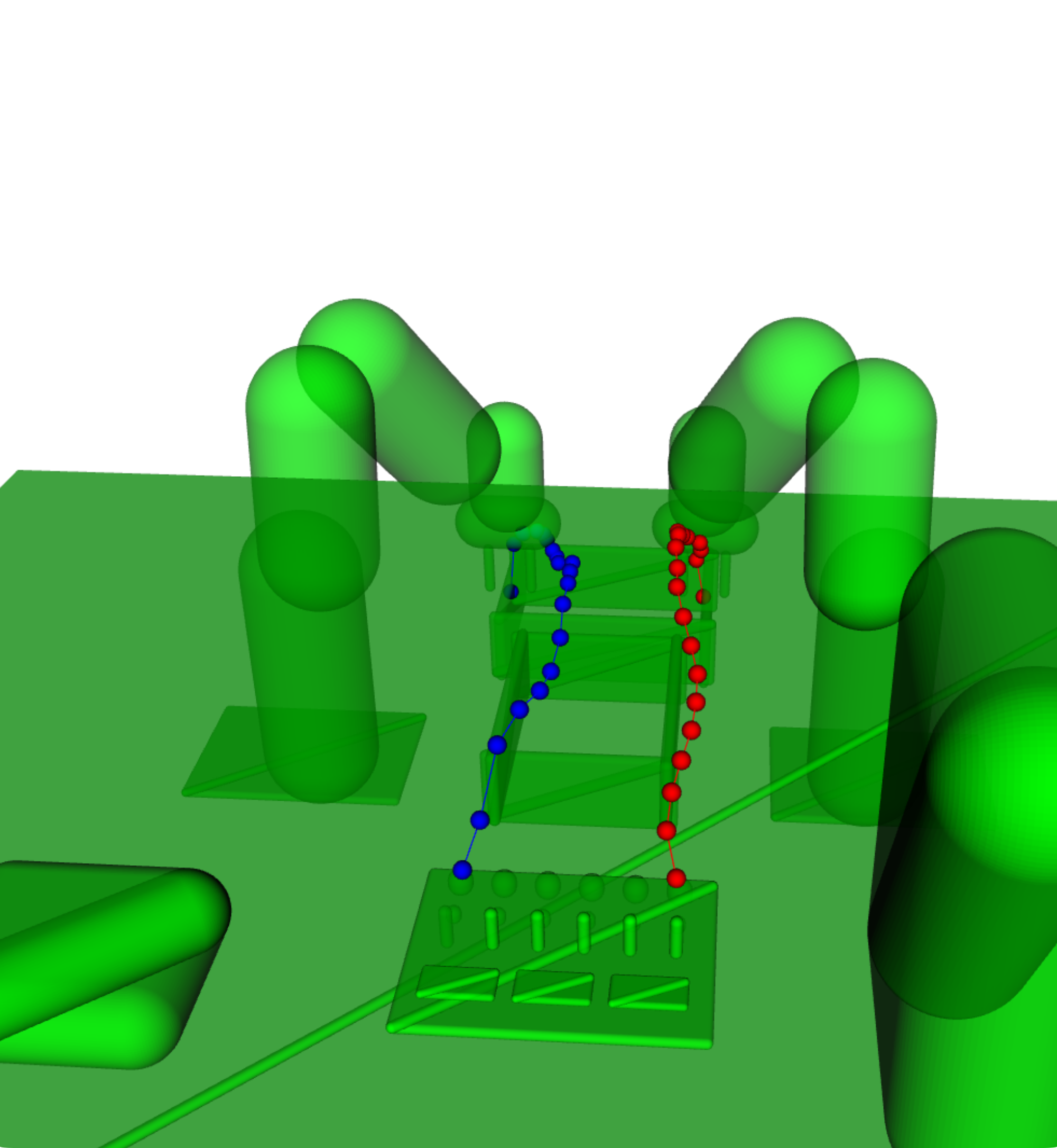}%
	}\hfil%
	\subfloat[The \textit{OMPL Path Simplifier} shortcuts the path to seven waypoints.\label{fig:OurShortcut}]{%
        \includegraphics[width=0.19\textwidth]{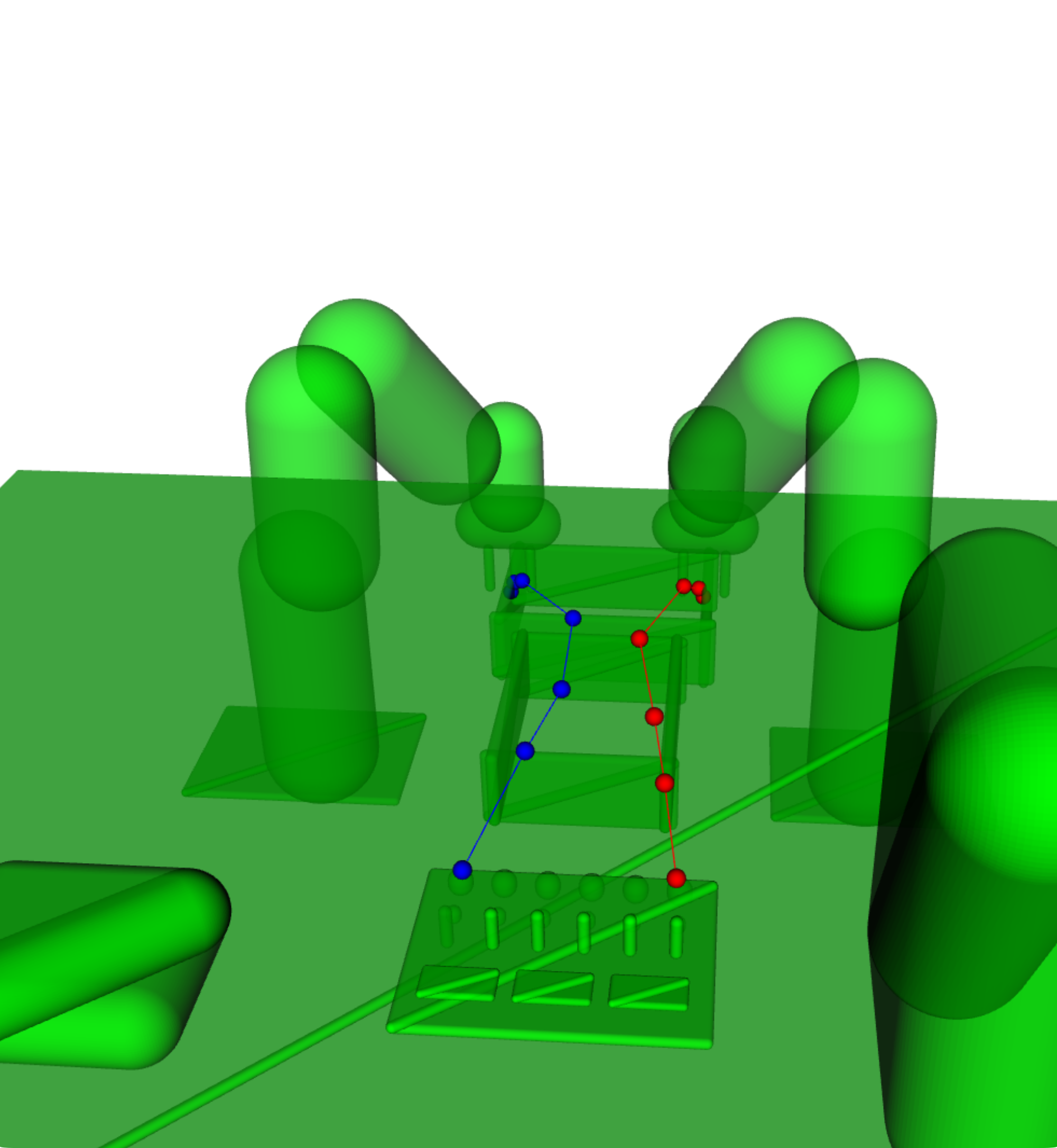}%
	}%
    \caption{Path planning steps and path post-processing steps exemplarily shown for the centralized approach. The spheres show the TCP positions for each configuration space waypoint computed with forward kinematics. Blue spheres are the TCP positions of the left robot, and red spheres are the TCP positions of the right robot.}\label{fig:PlanningPhases}%
\end{figure*}%
We performed the simulation study of Section~\ref{sec:Without} also using \textit{PLPP} and Table~\ref{table:WithPLPP} shows the results. The accompanying video at \href{https://youtu.be/6vaXqR7Gdew}{https://youtu.be/6vaXqR7Gdew} shows two example cycles, one for the centralized approach and one for the decoupled approach. As explained in Section~\ref{sec:PickAndPlace}, we only include the coordinated motion queries in this benchmark. For example, we do not include the cooperative manipulation of the blue box that places it from the table onto the mobile platform. Therefore, in this study and the video, we add the blue box twice: in its original pose on the table and its pose on the mobile platform after being picked up. For the centralized approach, Figure~\ref{fig:PlanningPhases} shows snapshots from the accompanying video at \href{https://youtu.be/6vaXqR7Gdew}{https://youtu.be/6vaXqR7Gdew} showing the path planning and path post-processing steps for one of the coordinated motion queries.\newline%
Table~\ref{table:WithPLPP} is similar to Table~\ref{table:WithoutPLPP}, but we add three bolded lines showing the computation time for \textit{PLPP} (line~4), the \textit{Original Path Length} before applying \textit{PLPP} (line~10), and the \textit{Modified Path Length} after applying \textit{PLPP} (line~11).\newline%
The \textit{Path Planner Times} are similar to the ones in Section~\ref{sec:Without}. The \textit{OMPL Path Simplifier Times} are higher than those given in Table~\ref{table:WithoutPLPP}, because now there is a second run of the \textit{OMPL Path Simplifier} as explained in Section~\ref{sec:Decoupled}, and we add the computation times of both runs. Also, the \textit{Trajectory Planner Times} are slightly higher than those in Table~\ref{table:WithoutPLPP} because we now align the paths closer to the obstacles, resulting in more path waypoints we need to interpolate.\newline%
Line~4 shows that the \textit{PLPP Time} is the dominant factor in the \textit{Motion Planning Time}, and in general, applying \textit{PLPP} significantly increases the \textit{Motion Planning Time}: For the centralized approach 0.26\,s without \textit{PLPP} vs. 2.73\,s with \textit{PLPP}, and for the decoupled approach 0.46\,s vs. 0.94\,s. Despite this increase, the \textit{Motion Planning Times} are still in the order of the \textit{Motion Durations}, so the approach with \textit{PLPP} is also suitable for online motion planning.\newline%
Distance computation is the main part of the \textit{Motion Planning Time} without \textit{PLPP}. With \textit{PLPP}, there is a second aspect that affects scalability: the SQP algorithm in \textit{PLPP}. Line~8 in Table~\ref{table:WithPLPP} shows that for the decoupled approach, distance computation still takes up most of the \textit{Motion Planning Time}. This is not the case for the centralized approach. Instead, the SQP algorithm we use scales poorly with the dimension of the problem. Note that we have included the distance computation times of the inequality constraints in the \textit{Distance Computation Time} of Table~\ref{table:WithPLPP}.\newline%
Lines 10-11 show that \textit{PLPP} significantly reduces the path length. For the centralized approach, the path length according to Eq.~(\ref{eq:PLPP_new}) is reduced by 77.2\,\%, and for the decoupled approach, the reduction is 54.6\,\%. For the centralized approach, this reduction in path length also reduces the \textit{Motion Duration}: 3.11\,s without \textit{PLPP} vs. 2.46\,s with \textit{PLPP}. For the decoupled approach, reducing the path length does not reduce the \textit{Motion Duration}. The reason is that in the decoupled approach, the separate minimization of the two paths results in two post-processed paths that suffer from many intersections, i.e. collisions. This requires many coordination motions of the fixed-path coordination, resulting in lower average velocities. Comparing Figure~\ref{fig:OMPLShorcut} with Figure~\ref{fig:OurShortcut}, the significant reduction in path length using \textit{PLPP} is obvious.\newline%
The total motion query time is the sum of the \textit{Motion Planning Time} and \textit{Motion Duration}. For the centralized approach, this gives 3.37\,s without \textit{PLPP} vs. 5.19\,s with \textit{PLPP}. For the decoupled approach, this gives 2.74\,s vs. 3.21\,s. We consider this increase acceptable, especially when considering the second benefit of \textit{PLPP} besides path length reduction: For both approaches, the shorter path lengths lead to fewer overshoots in the trajectory interpolation and thus to a significantly smaller number of \textit{Collision Failures} compared to the approach without \textit{PLPP}. For the decoupled approach, these higher quality paths also lead to significantly fewer \textit{Coordination Failures} than the approach without \textit{PLPP}. That is, the use \textit{PLPP} significantly increases the robustness.\newline%
Due to the smaller path lengths, which are more intuitive for human coworkers, and the significantly better robustness, we consider the increased computation time of \textit{PLPP} acceptable. With \textit{PLPP}, the centralized and decoupled approaches provide a motion planning pipeline that is acceptable in terms of robustness, computation time, and path quality (see Section~\ref{sec:Introduction}). The decoupled approach performs better in terms of \textit{Motion Planning Time} and total motion query time, which justifies the additional implementation effort. This is also shown in the accompanying video at \href{https://youtu.be/6vaXqR7Gdew}{https://youtu.be/6vaXqR7Gdew}, where the decoupled approach finishes the cycle with eight motion queries earlier than the centralized approach. We will use the decoupled approach with \textit{PLPP} in the following.\newline\newline%
Table~\ref{table:WithPLPP} shows a failure rate of $52 / 800 = 6.5\,\%$ for the centralized approach and a failure rate of 9.5\,\% for the decoupled approach. This is acceptable: In case of a planning failure, one can trigger the same motion query again, and the motion planning pipeline might compute a feasible solution on the second try. However, we point out that our multi-threaded implementation of the motion planning framework~\cite{Wittmann2023} allows us to easily run two instances of the same motion planning pipeline simultaneously without any additional effort. For example, we solved the same scenario of Table~\ref{table:WithPLPP} in a setup with three identical motion planning pipelines in parallel, i.e. all three used the decoupled approach with \textit{PLPP}. In this way, we achieved a robustness of 99.9\,\%, i.e. one failure among the 800 motion queries because for each of the 799 successful queries, at least one of the three motion planning pipelines computed a feasible trajectory. Solving the same motion query with multiple planners is not new. \cite{Raveh2011}~follows the same idea and combines the best parts of each solution path to increase the overall path quality.%
\subsubsection{Evaluation of our Implementation and Application to Hardware}%
Two key aspects make our implementation suitable for online motion planning: 1) we use \textit{SSV} modeling for distance computation instead of \textit{FCL} (see Section~\ref{sec:ThirdParty}); 2) we use the analytic gradients for the optimization in \textit{PLPP} that we derived in Section~\ref{sec:AnaGradients} and~\cite{Wittmann2022a}.\newline%
Table~\ref{table:Implemenation} shows the means and standard deviations of the \textit{Motion Planning Time} of different implementations for the scenario we also used in Sections~\ref{sec:Without} and~\ref{sec:With}. However, due to the longer computation times, we now average over five runs instead of 100 runs, i.e. we run 40 motion queries instead of 800. Instead of our \textit{SSV} modeling, one could use the \textit{FCL} modeling of \textit{MoveIt!} as shown in Figure~\ref{fig:fcl}. And instead of our analytically derived gradients, one could use numerically approximated gradients, and we use forward differences with a step size of $1\mathrm{e}{-6}$ in Table~\ref{table:Implemenation}.\newline%
It is obvious that only our \textit{SSV} modeling approach based on analytic gradients ensures motion planning times in the order of the motion duration, i.e. it is the only approach suitable for online motion planning.\newline\newline%
\begin{table}[ht]%
    \small%
    \centering%
    \caption{Path planning times for different implementations.}\label{table:Implemenation}%
    \begin{tabular}{@{}l|c@{}}%
        \toprule%
                                               & Motion Planning Time [s]\\%
        \midrule%
        \textit{SSV} with analytical grad.     & 1.19 $\pm$ 0.51\\%
        \textit{SSV} with numerical grad.      & 13.21 $\pm$ 8.92\\%
        \textit{FCL} with analytical grad.     & 14.32 $\pm$ 13.35\\%
        \textit{FCL} with numerical grad.      & 613.46 $\pm$ 693.23\\%
        \botrule%
    \end{tabular}%
\end{table}%
So far, we have given the results for simulated motions. However, the motions we compute are well applicable to hardware, and the accompanying video at \href{https://youtu.be/6vaXqR7Gdew}{https://youtu.be/6vaXqR7Gdew} shows an example of two cycles each with the eight motion queries as defined in Section~\ref{sec:Without} on our two \textit{Franka Emika} robots. We computed the motions using the decoupled approach with \textit{PLPP} and the multi-threaded approach outlined above with three motion planning pipeline instances running simultaneously. The average planning time before each motion is slightly longer than the 0.94\,s we give in Table~\ref{table:WithPLPP}. This is because in our current implementation, we wait for each motion planning pipeline to complete before executing the motion. The video only shows the motion of the cycle we defined to benchmark the approach. The real-world application with the other parts of the process, such as cooperative motions and grasping, is shown in our preliminary work~\cite{Wittmann2023} and at~\href{https://youtu.be/tH9bCa7sdIs}{https://youtu.be/tH9bCa7sdIs}.%
%
%
\section{Conclusions}\label{sec:Conclusions}%
Our main contribution in this work consists in a decoupled dual-arm planning approach that computes $\mathcal{C}^2$-continuous control commands. It is based on fixed-path coordination, a standard approach to coordinated planning in literature. We provide a detailed benchmark of a centralized and our decoupled approach. Thereby, for a standard motion planning pipeline such as the one of \textit{MoveIt!}, we show that the decoupled approach is inferior to centralized planning: It has a longer computation time because the additional coordination time outweighs the savings in path planning. And it has a lower robustness due to deadlocks.\newline%
Furthermore, another contribution of this work is an optimization problem with its gradients that minimizes rotational motions. It extends the existing path length minimization approach \textit{PLPP}. This contribution improves the path quality of the decoupled and centralized approaches, thus significantly increasing the robustness of both to over 90\,\%. However, the optimization does not scale well with the number of DoFs, so the centralized approach with \textit{PLPP} requires longer computation times than the decoupled approach. But both are still applicable to online motion planning, i.e. the motion planning times are below 3\,s.\newline%
With our multi-threaded implementation, we can further increase the robustness to up to 99.9\,\%. However, we still need to improve our implementation: Waiting for the longest planning pipeline before executing the motion increases the planning time, although it is still suitable for online planning.\newline%
We outline three limitations of our work: First, after the fixed-path coordination in the decoupled approach, we lack an additional feasibility check. We compute two trajectories separately, considering the maximum joint velocities, accelerations, and jerks. However, the subsequent coordination changes the dynamics, i.e. the velocities, accelerations, and jerks, of both trajectories. That is, one has to check again for violations. However, due to the nature of the fixed-path coordination, the velocity is always scaled down, so the check is only required for acceleration and jerk. Second, \textit{PLPP} only considers distances in Cartesian space. This could lead to long, unwanted configuration space motions near singular configurations. To avoid this, the optimization objective could include the configuration space distance. We provide the formula and the implementation in our preliminary work~\cite{Wittmann2022a}. Since we did not suffer from this problem in our scenario, we neglected configuration space distances in this work. Third, our \textit{SSV} modeling approach is computationally efficient compared to other approaches such as~\textit{FCL}. However, it provides only three types of simplified geometry: spheres, cylinders with rounded shape, and triangular volumes with rounded shape~\cite{broccoli}. For example, Figure~\ref{fig:ssv} shows that we model objects of rectangular shape with two triangular volumes. This modeling approach has limitations, e.g. in grasping scenarios, where the fingers of the end-effector must come into contact with gears. Figure~\ref{fig:ssv} shows that we model the gears, which have a cylindrical shape, with small spheres in the \textit{SSV} approach. This mismatch between the environment model and the real world can lead to collisions. Thus, the \textit{SSV} modeling is appropriate when the robot travels long distances in its workspace, where the objective is to avoid collisions. For assembly/contact-rich situations, one might prefer another modeling approach that can better represent reality.\newline%
When not using \textit{PLPP}, but instead using a standard motion planning pipeline such as in \textit{MoveIt!}, we show that the general assumption that decoupled planning is superior to centralized planning does not hold. However, it would be interesting for future work to investigate the scalability of planning time w.r.t. the number of robots. In our work, we focus only on dual-arm manipulation. Furthermore, there are open research areas in multi-robot planning that we do not cover in our work. For example, we did not answer the question of how to optimally assign tasks to the robots, i.e. the \textit{task assignment problem}.%
%
%
%
\backmatter
%
%
%
%
\section*{Statements and Declarations}
%
\textbf{Funding}\quad This work was supported by \textit{KME – Kompetenzzentrum Mittelstand GmbH}.\newline%
\textbf{Competing Interests - }The authors have no relevant financial or non-financial interests to disclose.\newline%
\textbf{Author Contributions - }\textbf{Jonas Wittmann}: Conceptualization, Methodology, Software, Validation, Formal Analysis, Investigation, Data Curation, Writing - Original Draft, Writing - Review and Editing, Visualization, Supervision, Project Administration, Funding Acquisition. \textbf{Franziska Ochsenfarth}: Methodology, Software, Validation, Investigation, Data Curation, Writing - Review and Editing. \textbf{Valentin Sonneville}: Methodology, Writing - Review and Editing. \textbf{Daniel Rixen:} Resources, Writing - Review \& Editing, Supervision, Project Administration, Funding Acquisition.\newline%
\textbf{Ethics approval - }Not applicable.\newline%
\textbf{Consent to participate - }Not applicable.\newline%
\textbf{Consent to publish - }Not applicable.\newline%
%
%
%
\bibliography{sn-bibliography}
\end{document}